\pdfoutput=1

\documentclass[11pt]{article}

\usepackage[]{acl}

\usepackage{times}
\usepackage{latexsym}

\usepackage[T1]{fontenc}

\usepackage[utf8]{inputenc}

\usepackage{microtype}

\usepackage{inconsolata}

\usepackage{graphicx}

\usepackage{xcolor}
\definecolor{mybrown}{RGB}{165, 42, 42}
\definecolor{myblue}{RGB}{165, 42, 42}
\definecolor{mygreen}{RGB}{5, 71, 42}

\usepackage{float}
\title{Exploring the Effectiveness and Consistency of Task Selection in Intermediate-Task Transfer Learning}%

\author{Pin-Jie Lin\textsuperscript{1} \,
 Miaoran Zhang\textsuperscript{2} \, Marius Mosbach\textsuperscript{3,4} \, \textbf{Dietrich Klakow}\textsuperscript{2} \\
\textsuperscript{1}Virginia Tech
\textsuperscript{2}Saarland University, Saarland Informatic Campus \\
\textsuperscript{3}Mila Quebec AI Institute 
\textsuperscript{4}McGill University\\
{\tt  pinjie@vt.edu}
}

\begin{document}
\maketitle

\begin{abstract}

Identifying beneficial tasks to transfer from is a critical step toward successful intermediate-task transfer learning. In this work, we experiment with 130 source-target task combinations and demonstrate that the transfer performance exhibits severe variance across different source tasks and training seeds, highlighting the crucial role of intermediate-task selection in a broader context. We compare four representative task selection methods in a unified setup, focusing on their effectiveness and consistency.
Compared to embedding-free methods and text embeddings, task embeddings constructed from fine-tuned weights can better estimate task transferability by improving task prediction scores from 2.59\% to 3.96\%.
Despite their strong performance, we observe that the task embeddings do not consistently demonstrate superiority for tasks requiring reasoning abilities.
Furthermore, we introduce a novel method that measures pairwise token similarity using maximum inner product search, leading to the highest performance in task prediction.
Our findings suggest that token-wise similarity is better predictive for predicting transferability compared to averaging weights.\footnote{We release the code publicly at \href{https://github.com/uds-lsv/intermediate-task-selection}{https://github.com/uds-lsv/intermediate-task-selection}.}
\end{abstract}

\section{Introduction}

Pre-trained language models (PLMs) have become foundational in the transfer learning paradigm of natural language processing (NLP) \cite{devlin-etal-2019-bert,NEURIPS2020_1457c0d6,JMLR:v24:22-1144}.
Intermediate-task transfer learning aims to improve model performance further by introducing an intermediate stage of supervised training on data-rich tasks before fine-tuning the target downstream task \cite{DBLP:journals/corr/abs-1811-01088, pruksachatkun-etal-2020-intermediate, vu-etal-2020-exploring}.
The paradigm has shown to be particularly useful for improving performance in resource-constrained scenarios where annotated training data is often limited \cite{prasad-etal-2021-effectiveness,vu-etal-2022-spot}.

\begin{figure}[t]
    \centering
    \includegraphics[width=0.5\textwidth]{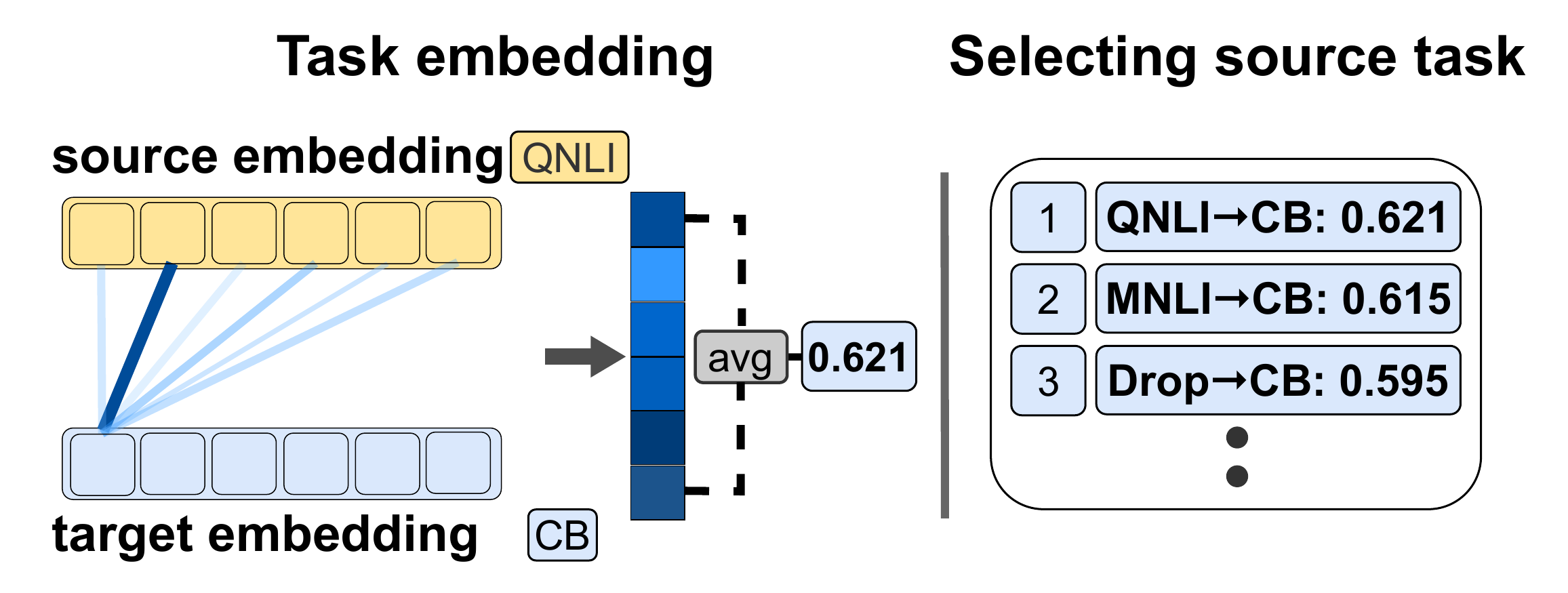}
    \caption{
    Our proposed method, \emph{maximum inner product search}, is based on pairwise token similarity. Left: Given a target task (e.g., \ssc{CB}), we obtain the maximum token-wise similarity scores between the target and the source tasks for each embedding position. Right: We select the source task with the highest mean of maximum similarity scores.}
    \label{fig:max_task_embedding}
    \vspace{-5mm}
\end{figure}

A crucial aspect of intermediate-task transfer learning is to select beneficial tasks to transfer from. However, the costs of searching for the optimal intermediate-task, especially with the growing array of available NLP tasks and the exhaustive process of model fine-tuning 
\cite{pruksachatkun-etal-2020-intermediate,vu-etal-2020-exploring}, are prohibitive.
Research on intermediate-task selection mainly predicts task transferability using task-specific embeddings, which condense the task information of a given target task into a single vector representation. For example, some works construct task embedding from fine-tuned weights~\cite{vu-etal-2022-spot,zhou-etal-2022-efficiently} or leverage text embedding~\cite{poth-etal-2021-pre}. More specifically, 
\citet{poth-etal-2021-pre} use sentence transformers to encode dataset examples as text embeddings.
The more recent approach by \citet{vu-etal-2022-spot} constructs task embeddings from the weights of soft prompts, which have been effectively applied in large-scale studies.

Despite their promising results, a systematic study of the consistency of these task selection methods is still missing.
Specifically, it remains unclear how consistent these approaches are at predicting the best source task to transfer from. To address this gap, we perform a comprehensive evaluation of existing task selection methods in intermediate-task transfer learning. 
Our research questions are:
(1) \emph{Do intermediate-task selection approaches exhibit consistent performance across downstream tasks?}
(2) \emph{What are the key ingredients that result in accurate transferability predictions?} %

To answer these questions, we perform experiments across 130 intermediate and downstream task combinations derived from 13 source and 10 target tasks. Our results show that intermediate-task transfer exhibits significant performance variance across tasks. Comparing four representative task selection methods, we find that task embeddings based on fine-tuned weights \citep{vu-etal-2022-spot} generally outperform embedding-free and text embedding methods \cite{poth-etal-2021-pre}.
However, we also observe that such task embeddings do not consistently perform well on tasks requiring high-level reasoning abilities.
Exploring this further, we revisit the task embedding design and propose a new construction method based on pairwise token similarity (see Figure \ref{fig:max_task_embedding}), which yields the highest average task prediction performance of 82.5\%. 
Our main contributions are as follows:
\begin{enumerate}
    \item We systematically investigate intermediate-task transfer learning across 130 intermediate and downstream task combinations.
    \item We examine four representative task selection methods in a unified setup, including both embedding-free and embedding-based methods.
    \item We introduce a novel task embedding construction approach based on pairwise token similarity, which achieves the highest task prediction performance of 82.5\% in nDCG score.
    \item We provide an in-depth analysis of the impact of task type and training seed, along with an exploration into embedding distributions.
\end{enumerate}

\begin{figure*}[th!]
    \includegraphics[width=1.0\textwidth,height=1.0\textheight,keepaspectratio]{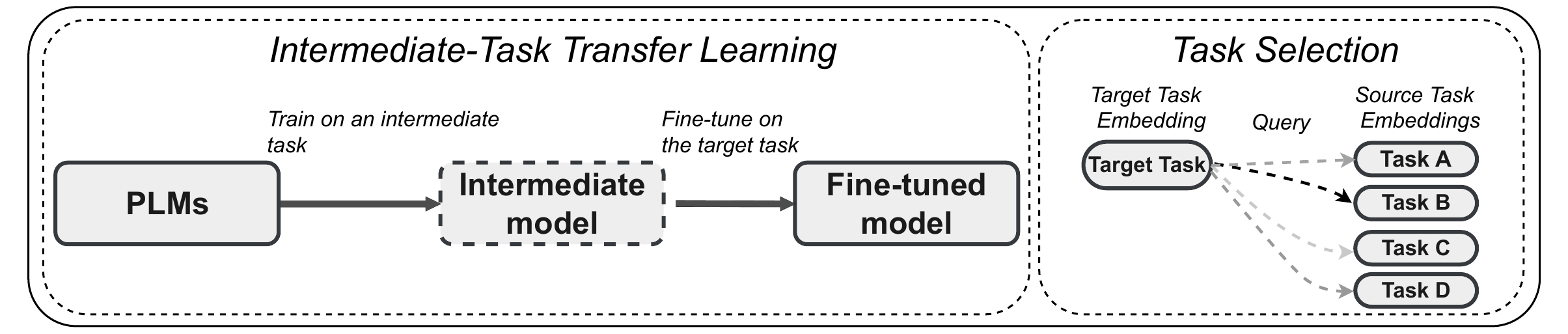}
    \caption{Left: \textbf{Intermediate-task transfer learning} performs sequentially learning on the source task followed by fine-tuning on the target task. Right: \textbf{Task selection} is a process where given a target task, the goal is to identify the most beneficial task for transfer by searching over a set of source tasks through its task embedding. The selection process relies on a similarity metric to measure the transferability of tasks or datasets. \label{fig:iit_main_overview}}
    \vspace{-3mm}
\end{figure*}

\section{Related Work}
\label{section_2}

Identifying a beneficial task from a broader set of source tasks is a crucial step in intermediate-task transfer learning. 
Various studies have proposed methods to estimate task transferability based on task embeddings.

A foundational approach is Task2Vec \cite{DBLP:journals/corr/abs-1902-03545,vu-etal-2020-exploring}, which involves computing the Fisher information matrix and enables to measure semantic and taxonomic relationships between tasks. In contrast, \citet{poth-etal-2021-pre} demonstrate the effectiveness of text embeddings based on sentence encoders.
The landscape of task selection approaches has further evolved with the introduction of parameter-efficient fine-tuning (PEFT) techniques. 
For instance, \citet{vu-etal-2022-spot} use soft prompts to generate task embeddings, demonstrating the effectiveness of prompt-based embeddings. 
Expanding on this, \citet{zhou-etal-2022-efficiently} investigate other PEFT methods, including P-tuning \cite{liu-etal-2022-p,liu2022ptuning}, fine-tuning only bias terms \cite{ben-zaken-etal-2022-bitfit}, and LoRA \cite{hu2022lora}. They construct task embeddings based on the fine-tuned weights.

Task selection based on neuron activations provides another perspective by focusing on the patterns of activations within models. \citet{su-etal-2022-transferability} propose model stimulation similarity to identify beneficial source tasks through the overlap rate of activations. More recently, \citet{xi-etal-2023-connectivity} introduce connectivity patterns as task embeddings, identifying task-specific patterns in deep neural networks that best represent the tasks.

Our work differs from previous studies by contributing a comparison of existing task selection methods in a unified setup, specifically focusing on the effectiveness and consistency of these approaches.

\section{Background}
\label{sec:section_3}

In the following, we introduce the intermediate-task transfer learning paradigm and motivate our focus on parameter-efficient fine-tuning.

\subsection{Intermediate-Task Transfer Learning}

As depicted in Figure \ref{fig:iit_main_overview}, intermediate-task training involves sequentially fine-tuning on a source task followed by fine-tuning on a target task. 
By incorporating an intermediate stage of supervision (typically on data-rich tasks), intermediate-task transfer learning enables knowledge transfer across tasks, thereby enhancing performance on low-resource target tasks \cite{vu-etal-2022-spot}.

More formally, the intermediate-task transfer learning paradigm can be divided into two stages: (1) training a PLM $f_{\theta}$ on a given source task $\bmmc{T}_{s}$ to obtain the intermediate model $f_{\theta^{'}}$; (2) training the intermediate model $f_{\theta^{'}}$ on the target task $\bmmc{T}_{t}$. 
The objective function with a cross-entropy loss $\bmmc{L}$ of the first stage is defined as follows:
\begin{align}
   \bmmc{\theta}^{'} = \argmin_{\bmmc{\theta}} \bmmc{L}_{\bmmc{T}_{s}} (f_{\theta}).
\end{align}

Here, the source task $\bmmc{T}_{s}$ is selected based on a selection criterion using metadata of datasets, domain similarity, or task similarity.  
Subsequently, the intermediate model is trained on the target task:

\begin{equation}
    \label{eq:target}
    \bmmc{\theta}^* = \argmin_{\bmmc{\theta}^{'}} \bmmc{L}_{\bmmc{T}_{t}}(f_{\theta^{'}})
\end{equation}

Note that in Equation \ref{eq:target} the intermediate model $f$ is parameterized with $\bmmc{\theta}^{'}$, representing the parameters of the model trained on source task $\bmmc{T}_{s}$.

\subsection{Parameter-Efficient Fine-Tuning via Soft Prompts}
Modern language models often contain billions of parameters, making sequential fine-tuning and experimenting with a large number of source and target task combinations impractical.
Recent studies have explored parameter-efficient fine-tuning approach through prompt tuning, which involves learning task-specific soft prompts that allow a frozen language model to efficiently perform specific downstream tasks~\cite{lester-etal-2021-power,li-liang-2021-prefix,liu-etal-2022-p}.
Unlike discrete prompts, soft prompts consist of a set of learnable prompt tokens that are learned through backpropagation and can be applied to various downstream tasks. 
This approach has been successfully used to efficiently adapt large language models in various scenarios~\cite{qin-eisner-2021-learning,vu-etal-2022-overcoming,asai-etal-2022-attempt}.More recently, researchers have focused on intermediate-task transfer learning using prompt tuning, specifically Soft Prompt Transfer (SPoT) \cite{vu-etal-2022-spot}. SPoT employs a series of soft prompt tokens to adapt frozen models to specific downstream tasks, making it highly parameter-efficient for intermediate-task transfer learning. In this transfer learning procedure, a pre-trained model is adapted to each task by conditioning on a set of learnable prompt tokens. Moreover, the resulting prompts can directly serve as task embeddings to assess task transferability.

\section{Intermediate-Task Selection Methods}
\label{sec:identifying_intermediate_task}

\begin{table}[t]
\centering
\begin{adjustbox}{max width=0.48\textwidth}
\begin{tabular}{ l c c c}
\toprule
\textbf{Method} & \bssc{Dataset $D$}  & \bssc{Model $f$} & \bssc{Output} \\
\midrule
\multicolumn{4}{c}{\cellcolor{gray!25}\ssc{Embedding-free}} \\
\midrule
\midrule
\hspace{5pt}\ssc{Random} & \xmark & \xmark & - \\ 
\midrule
\multicolumn{4}{l}{\ssc{Metadata}} \\ %
\hspace{5pt}\ssc{Size} & \cmark & \xmark & $\mathbb{R}$ \\ 
\midrule
\midrule
\multicolumn{4}{c}{\cellcolor{gray!25}\ssc{Embedding-based}} \\
\midrule
\multicolumn{4}{l}{\ssc{Text embedding}} \\
\hspace{5pt}\ssc{SEmb} & \cmark& \cmark & $\mathbb{R}^{d}$ \\ 
\midrule
\midrule
\multicolumn{4}{l}{\ssc{Task embedding}} \\
\hspace{5pt}\ssc{Feature} & \cmark & \cmark &  $\mathbb{R}^{d}$ \\ 
\midrule
\end{tabular}

\end{adjustbox}
\caption{
An overview of task selection methods. These task selection methods differ in whether the dataset 
$D$ and a model $f$ is used for selection and their output format.
Note that \bssc{SEmb} relies on sentence encoder models, while \bssc{Feature} requires intermediate models to construct task embeddings.} \label{tbl:overview_task_selection}
\vspace{-3mm}
\end{table}

Intermediate-task transfer can improve the performance of the target downstream task, but it is computationally infeasible to try out all possible task combinations, making choosing a beneficial source task an important problem.

\textit{Intermediate-task selection} aims to predict task transferability and retrieve the most beneficial task from a broad set of available source tasks. This eliminates the need for exhaustive training and is more feasible in resource-constrained scenarios.
Here, we compare existing intermediate-task selection methods which can be categorized into two groups: \emph{embedding-free} and \emph{embedding-based} methods (see Table \ref{tbl:overview_task_selection}).

\subsection{Embedding-Free Methods} 

The first group of methods operates without accessing any model.
They estimate task transferability based on certain criteria, such as data size, or simply perform random selection. These methods serve as baseline approaches in \citet{poth-etal-2021-pre}.

\paragraph{Random selection (\bssc{Random})} This method selects the intermediate-tasks randomly without using any specific information for the tasks and models.

\paragraph{Data size (\bssc{Size})} This method predicts the task transferability based on the data size, assuming that larger datasets indicate higher transferability to model performance.

\subsection{Embedding Methods}

The second group of methods constructs embeddings either using a pre-trained sentence encoder model or an intermediate model $f_{\theta^{'}}$.
We consider two such methods:

\paragraph{Sentence embeddings (\bssc{SEmb})}
It represents the text embedding obtained by averaging all sentence representations on the whole dataset~\cite{poth-etal-2021-pre}.
Each sentence representation, denoted as $h_{x_i}$, is encoded by the encoder model for a given example $x_i$.
These sentence representations are averaged over the entire dataset: $\sum_{x_{i}\sim{\bmmc{D}}} \frac{h_{x_i}}{|\bmmc{D}|}$. 
This method captures linguistic properties of the input text $x$ for both the source and target tasks, independent of the intermediate-task training algorithm.

\paragraph{Prompt similarity (\bssc{Feature})}
It measures task similarity based on the similarity between their task-specific prompts and employs solely fine-tuned weights to create task embeddings \cite{vu-etal-2022-spot}.
Let the prompt weights be denoted as $[e_1, e_2, ...e_{N}]\in \mathbb{R}^{N\times d}$, consisting of $N$ soft prompt tokens with $d$ feature dimensions.
 The prompt similarity score between two tasks, $t^{1}$ and $t^{2}$, is defined as the cosine similarity of the average representations of prompt tokens:
\begin{equation}
    {\rm sim}(t^{1}, t^{2}) = \cos(\frac{1}{N} \sum_{i=1}^{N} e_{i}^{1}, \frac{1}{N} \sum_{j=1}^{N} e_{j}^{2})
\end{equation}
where $e_{i}^{1}$ and $e_{j}^{2}$ represent the prompt token representations of the tasks $t^{1}$ and $t^{2}$, and $\cos$ denotes the cosine similarity.
This method computes the task embedding, represented as a vector in $\mathbb{R}^{d}$, by averaging the feature values across all prompt tokens.
We refer to this method as \bssc{Feature} to emphasize its focus on capturing task-specific features.

\begin{table}[t]
\centering
\scalebox{0.9}{
\begin{tabular}{l*{3}{S[table-format=6.2,table-number-alignment=left]}}
\toprule
\textbf{Name}	&	\textbf{Task}	&	\textbf{|Train|} \\	
\midrule
\emph{source tasks}	\\	
\hspace{2mm} \ssc{MNLI} &	\text{NLI}	&	\text{393K}	\\	
\hspace{2mm} \ssc{QQP} &	\text{paragraph detection}	&	\text{364K}	\\	
\hspace{2mm} \ssc{QNLI} &	\text{NLI}	&	\text{105K}	\\	

\hspace{2mm} \ssc{ReCoRD} &	\text{QA}	&	\text{101K}	\\	
\hspace{2mm} \ssc{CXC} &	\text{semantic similarity}	&	\text{88K}	\\	
\hspace{2mm} \ssc{SQuAD} &	\text{QA}	&	\text{88K}	\\	

\hspace{2mm} \ssc{DROP} &	\text{QA}	&	\text{77K}	\\	
\hspace{2mm} \ssc{SST-2} &	\text{sentiment analysis}	&	\text{67K}	\\	
\hspace{2mm} \ssc{WinoGrande} &	\text{commonsense reasoning}	&	\text{40K}	\\	

\hspace{2mm} \ssc{HellaSWAG} &	\text{commonsense reasoning}	&	\text{40K}	\\	
\hspace{2mm} \ssc{MultiRC} &	\text{QA}	&	\text{27K}	\\	
\hspace{2mm} \ssc{CosmosQA} &	\text{commonsense reasoning}	&	\text{25K}	\\	

\hspace{2mm} \ssc{RACE} &	\text{QA}	&	\text{25K}	\\	
\midrule
\midrule
\emph{target tasks}	\\
\hspace{2mm} \ssc{BoolQ}	& QA & \text{9K}	\\	
\hspace{2mm} \ssc{CoLA} & \text{grammatical acceptability} & \text{9K}	\\
\hspace{2mm} \ssc{STS-B} & \text{semantic similarity} & \text{6K}	\\

\hspace{2mm} \ssc{WiC} & \text{word sense disambiguation} & \text{5K} \\
\hspace{2mm} \ssc{CR} & \text{sentiment analysis} & \text{4K}	\\
\hspace{2mm} \ssc{MRPC} & \text{paraphrase detection} & \text{4K} \\	

\hspace{2mm} \ssc{RTE} &	\text{NLI}	& \text{2K} \\	
\hspace{2mm} \ssc{WSC} & \text{coreference resolution} & \text{554}	\\
\hspace{2mm} \ssc{COPA} & \text{QA} & \text{400}	\\

\hspace{2mm} \ssc{CB} & \text{NLI} & \text{250}	\\
\hline
\end{tabular}}
\caption{Overview of source and target tasks. For intermediate-task transfer, we first train on one of the source tasks and then continually fine-tune on the target task. 
\label{tab:dataset}}
\vspace{-3mm}
\end{table}

\section{Systematic Evaluation of Task Selection Methods}
\label{sec:results}

\subsection{Experimental Setup}

\paragraph{Datasets.} 
We consider 13 source tasks of various types, including question answering (QA), natural language inference (NLI), and sentiment analysis, among others. We evaluate the transfer performance on 10 target tasks, following the setting in \citet{vu-etal-2022-spot}, as presented in Table \ref{tab:dataset}.
More details on the datasets are provided in Appendix \ref{app:detailed_dataset}.

\paragraph{Models.}

For all experiments, we adopt \ssc{T5 Base} \cite{JMLR:v21:20-074} as our PLM. The pre-trained weights remain frozen, and only the weights of the soft prompt tokens are updated. After training, these fine-tuned weights are then used to construct task embeddings and perform soft prompt transfer.\looseness-1

\paragraph{Implementation details.}  We closely follow the training configurations outlined in \citet{lester-etal-2021-power}. We train soft prompts for 30K steps, using three random seeds (42, 150, 386).
We use $N=100$ prompt tokens and initialize the weights of the prompt tokens from the embeddings of the top 5K most frequent tokens in the pre-training data. We use the AdaFactor optimizer \cite{pmlr-v80-shazeer18a} with a linear scheduler. 
After conducting prompt tuning, we select the best-performing checkpoint for prompt transfer. The prompt transfer experiment is conducted with another set of training seeds (112, 28, 52)
effectiveness of prompt transfer using a relative transfer performance metric, calculated as follows: $\frac{M_{s \rightarrow t} - M_{t}}{M_{t}}$.
Here, the $M_{t}$ indicates the model performance with no-transfer prompt tuning, and $M_{s \rightarrow t}$ represents the transfer performance. The evaluation metric for the model performance varies according to individual tasks.

\subsection{Task Selection Methods and Evaluation}

\paragraph{Embedding-based methods.}
For text embeddings, we follow the model choice in \citet{poth-etal-2021-pre}.
We use the off-the-shelf encoder models to derive sentence representations for both source and target tasks. 
Specifically, we adopt Sentence-BERT and Sentence-RoBERTa~\cite{reimers-gurevych-2019-sentence} as encoders for \bssc{SEmb-B} and \bssc{SEmb-R}, respectively. \looseness-1

\paragraph{Selection criterion.}
We rank the order of beneficial tasks based on quantitative values from embedding-free methods.
For embedding-based methods on tasks $t^{1}$ and $t^{2}$, we employ cosine similarity using the mapping function $h(\cdot)$  to construct the task embedding or text embedding for a given intermediate task.
To get the ranking order, we sort the source tasks based on the score ${\rm sim}(t^{1}, t^{2}) = \cos(h(t^{1}), h(t^{2}))$ between the source and target tasks. The ground-truth ranking is obtained by transferring source tasks to the downstream task and sorting them based on transfer performance. \looseness-1 \raggedcolumns

\paragraph{Evaluation.} \label{metrics}

We use two metrics\footnote{See formal definitions in Appendix~\ref{app:metrics}.} to evaluate the effectiveness of task selection methods: (1) Normalized Discounted Cumulative Gain (nDCG) \cite{10.1145/582415.582418}, 
a widely accepted information retrieval measure that evaluates the overall quality of a ranking, emphasizing the entire order rather than merely focusing on the rank of the best source task. The nDCG score ranges from 0 to 1, where 1 presents the exact match with the ideal order and lower values indicate a lower quality of ranking.
(2) Regret@k \cite{Renggli_2022_CVPR}, a metric for computational regret, quantifying the relative performance between the expected performance of the top-$k$ selected intermediate-tasks and the optimal intermediate-task.
Lower regret signifies a more effective selection strategy among the $k$ intermediate models. 
For each target task, we evaluate the overall ranking prediction of the 13 source tasks against the ground-truth ranking using nDCG score.
We evaluate the efficacy of the top-$k$ selected source tasks compared to the ground-truth selection using Regret@k.

\subsection{Results}\label{section_prompt_transfer}
\begin{figure}[t]
    \centering
    \includegraphics[width=0.49\textwidth]{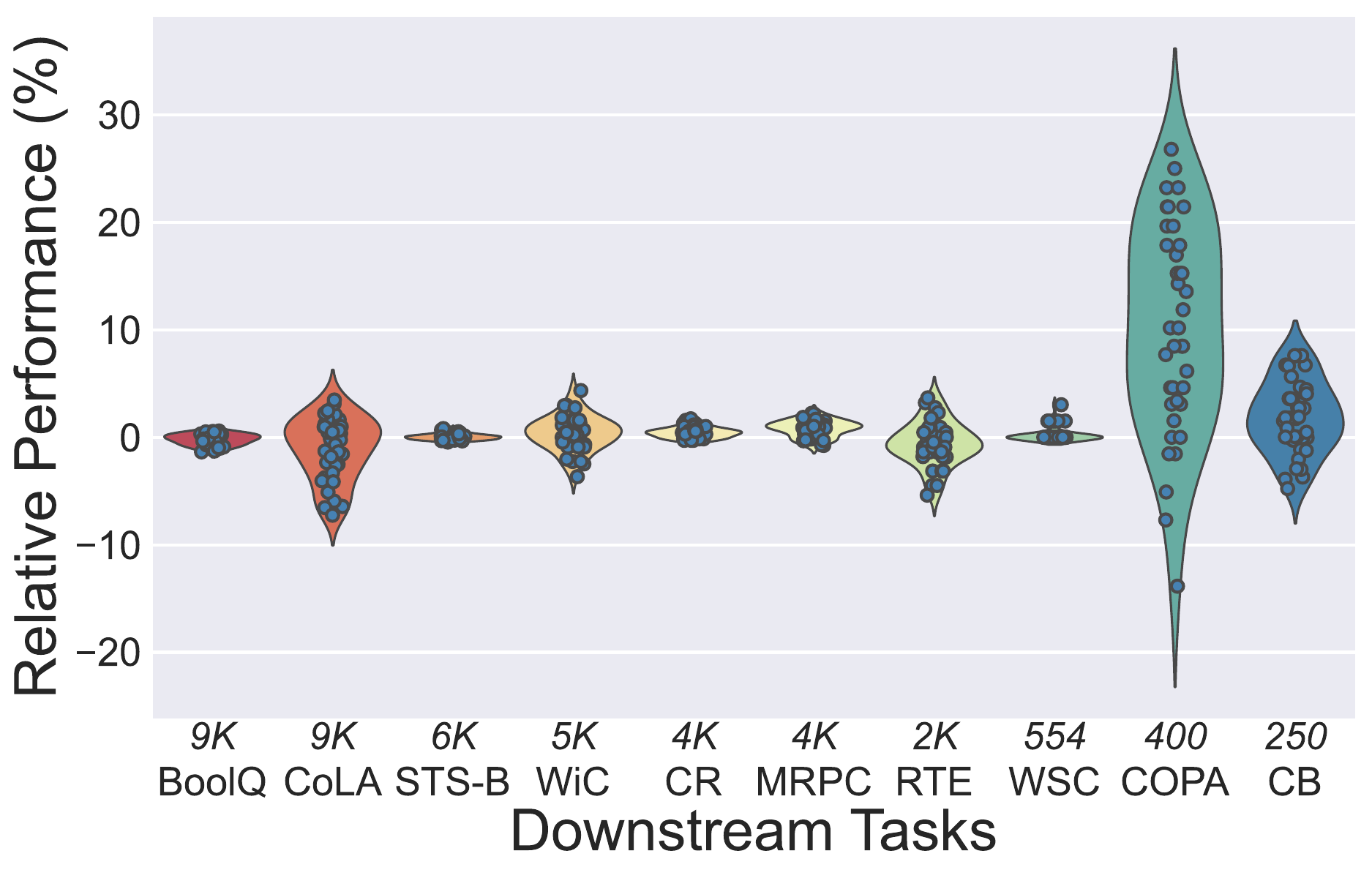}
    \caption{
    Relative transfer performance across ten downstream tasks with 390 intermediate-task trained models (13 source $\times$ 10 target tasks $\times$ 3 seeds).   
    Each violin plot illustrates the distribution of performance on the x-axis, with each dot denoting the relative improvement or deterioration compared to the no-transfer baseline on the y-axis.
    Tasks are arranged in descending order of the training sample sizes.
    }\label{fig:transfer_gain_pct}
    \vspace{-3mm}
\end{figure}

\begin{table*}[th!]
\begin{subtable}[t]{\linewidth}
    \centering
    \def\arraystretch{0.8}
    \begin{tabular}{lrrrrrr|rr}
\toprule
{} & \multicolumn{2}{c}{\bssc{Classification}} & \multicolumn{2}{c}{\bssc{M. Choice}} & \multicolumn{2}{c}{\bssc{QA}} & \multicolumn{2}{c}{\bssc{All}} \\
 \cmidrule(lr){2-3} \cmidrule(lr){4-5} \cmidrule(lr){6-7} \cmidrule(lr){8-9} 

{} & R@1$\downarrow$ & nDCG$\uparrow$ & R@1$\downarrow$ & nDCG$\uparrow$ & R@1$\downarrow$ & nDCG$\uparrow$ & R@1$\downarrow$ & nDCG$\uparrow$ \\
\cmidrule(lr){2-9}

\bssc{Random} & 2.18 & 81.53 & 2.20 & 84.52  & 1.45 & 86.43 & 2.89 & 77.89 \\
\cmidrule(lr){2-9}

\bssc{Size} & 2.10  & 83.73 & \textbf{1.44} & 86.01 & \textbf{0.88} & 90.06 & 2.78 & 78.00 \\

\cmidrule(lr){2-9}
\bssc{SEmb-B} & 1.92  & 85.21 & 1.91 & 86.12 & 1.21 & 90.11 & 2.75 & 78.23 \\
\bssc{SEmb-R} & 1.82  & 86.51 & 1.74 & 86.31 & {1.12} & 90.23 & 2.32 & 79.26 \\

\cmidrule(lr){2-9}
\bssc{Feature} & \textbf{1.28} & \textbf{87.31} & 1.67 & \textbf{86.40} & {1.02} & \textbf{90.70} & \textbf{2.04} & \textbf{81.85}  \\
\bottomrule
\end{tabular}
    \end{subtable}
        \caption{Comparison of task selection methods on 10 downstream tasks. 
        The nDCG and Regret@1 (R@1) scores are grouped by the target task category and we report the mean scores for each group. 
        The best scores in each group are boldfaced.}\label{tab:method_results}
        \vspace{-3mm}
\end{table*}

\paragraph{Intermediate-task transfer exhibits high-performance variance across tasks.}
Figure~\ref{fig:transfer_gain_pct} illustrates the relative transfer performance across 10 target tasks, sorted by their training data sizes~\footnote{The detailed transfer performances are presented in Appendix~\ref{appendix:emprical_prompt_transfer_performance}.}.
We find that relative transfer performance through intermediate-task training exhibits significant variance across tasks, especially for the downstream tasks \ssc{CoLA}, \ssc{RTE}, \ssc{COPA}, and \ssc{CB}. This observation aligns with previous studies showing significant performance variation across source tasks~\cite{pruksachatkun-etal-2020-intermediate,jiang2023forkmerge}. Additionally, we find that this phenomenon is particularly pronounced in downstream tasks with extremely limited labeled data, such as \ssc{COPA} and \ssc{CB}. In contrast, the relative transfer performance is more consistent for downstream tasks that have sufficient training data, like \ssc{BoolQ} and \ssc{STS-B}. In Appendix~\ref{appendix:transfer_gain_with_data_size}, we show that there exists a correlation between transfer gains and training data sizes. These results highlight the importance of carefully selecting beneficial tasks to enhance transfer gains, especially in low-resource scenarios.

\paragraph{Embedding-based selection methods outperform embedding-free methods, but the transfer gains are limited.}

Table \ref{tab:method_results} presents results for the four task selection methods.
Embedding-based approaches show higher task prediction performance over embedding-free methods, indicating richer information is obtained from encoded representations for predicting task transferability.
Specifically, \bssc{Feature} outperforms all other task selection methods on average.
Despite its strong performance, \bssc{Feature} falls short of the simple \bssc{Size} approach in Regret@1 for multiple choice (\bssc{M. Choice}) and question answering (\bssc{QA}) tasks. This highlights the need to further improve task embeddings, especially for tasks that require reasoning abilities.
\begin{table}[ht]
    \centering    
    \begin{adjustbox}{max width=0.5\textwidth}
    \begin{tabular}{lrr|c}
    \toprule
    {} & \multicolumn{2}{c}{\bssc{Transfer Gain}} & \multirow{2}{*}{\bssc{Avg. score}} \\
    \cmidrule(lr){2-3} 
    {} & \bssc{Abs}. & \bssc{Rel.} &   \\
    \cmidrule(lr){1-4}
    \cmidrule(lr){3-4}
     \bssc{No Transfer} & - & \text{-} & \text{77.2}  \\
    \cmidrule(lr){1-4}
    \bssc{Random} & 0.38 & \text{0.49} & \text{77.58}  \\
    \bssc{Size} & 0.52 & \text{0.67} & \text{77.72}  \\
    \cmidrule(lr){1-4}
    \ssc{ \bssc{SEmb-R}}  &  &  &   \\
    \ssc{ Best of Top-k} & &  & \\
    \hspace*{+2mm} \ssc{ $k$=1} & 0.72 & \text{0.93} & \text{77.92}  \\
    \hspace*{+2mm} \ssc{ $k$=3} & \text{0.78} & \text{1.01} & \text{77.98}  \\
    \cmidrule(lr){1-4}
    \ssc{ \bssc{Feature}}  &  &  &   \\
    \ssc{ Best of Top-k} & &  & \\
    \hspace*{+2mm} \ssc{ $k$=1} & 0.91 & \text{1.17} & \text{78.11}  \\
    \hspace*{+2mm} \ssc{ $k$=3} & \textbf{1.03} & \textbf{1.33} & \textbf{78.23}  \\
    \bottomrule
    \end{tabular}
    \end{adjustbox}
    \caption{Comparison of task selection methods on model performance. \bssc{Abs} and \bssc{Rel} represent absolute and relative improvements compared to no-transfer baseline. \bssc{Avg. score} is calculated across 10 downstream tasks with three runs. \ssc{Best of Top-k} is the best performance across the top-$k$ selected source tasks.
    }\label{tab:task_emb_for_selection}
    \vspace{-3mm}
\end{table}

\begin{table*}[t!]
\centering
\begin{adjustbox}{max width=\textwidth}
\begin{tabular}{l|lll|lll|lll}
\toprule
\multirow{2}{*}{\bssc{Target}} & \multicolumn{3}{c|}{\bsscemph{{seed 112}}} & \multicolumn{3}{c|}{\bsscemph{{28}}} & \multicolumn{3}{c}{\bsscemph{{52}}} \\
&	\bssc{Source}	& \bssc{Task Type} & \bssc{Rel. (\%)} &	\bssc{Source}	& \bssc{Task Type} & \bssc{Rel. (\%)} & \bssc{Source}	& \bssc{Task Type} & \bssc{Rel. (\%)} \\
\midrule
\midrule

\emph{Top-3 transfer} & \multicolumn{3}{c|}{} & \multicolumn{3}{c|}{} & \multicolumn{3}{c}{} \\
\hspace{1mm} COPA (QA) & \text{MultiRC*} & \text{QA} & \text{7.69}  & \text{CxC} & \text{semantic sim.} & \text{16.94} & \text{QQP} &\text{paraphrase} & 26.78 \\	
\hspace{1mm}& \text{DROP*} & \text{QA} & \text{6.15}  & \text{MultiRC*/RACE*} & \text{QA/QA} & \text{15.25} & \text{ReCORD*} & \text{QA} & 24.99 \\
\hspace{1mm}& \text{RACE*} & \text{QA} & \text{4.61} & \text{QQP} &\text{paraphrase}& \text{13.55} & \text{WinoGr./MultiRC*} &\text{reasoning/QA} & 23.21 \\

\midrule
\midrule
\emph{Top-3 transfer} & \multicolumn{3}{c|}{} & \multicolumn{3}{c|}{} & \multicolumn{3}{c}{} \\
\hspace{1mm} CB (NLI) & \text{QNLI*} & \text{NLI} &  \text{4.11} & \text{RACE}	&\text{QA} & \text{4.04} & 
\text{CxC/RACE}	& \text{semantic sim./QA} & \text{7.60} \\	
\hspace{1mm}  &
\text{MNLI/WinoGr.} & \text{NLI/reasoning} & \text{3.61} & \text{ReCORD} & \text{QA} & \text{3.53} 
& \text{ReCORD} & \text{QA} &	\text{7.57} \\	
\hspace{1mm} &	
\text{SQuAD} &\text{QA} & \text{2.70} & \text{SQuAD} & \text{QA}& \text{2.73} &
\text{QNLI/HellaSWAG} & \text{NLI/reasoning} & \text{7.72} \\	
\bottomrule
\end{tabular}
\end{adjustbox}
\caption{Top-3 intermediate-task transfer on \ssc{COPA} and \ssc{CB}. \bssc{Rel.} is the relative performance improvement (\%) calculated based on the corresponding no-transfer prompt tuning. * indicates that the source task type is identical to the downstream task type.}
\label{tab:correlation_task_type} 
\vspace{-3mm}
\end{table*}

In Table~\ref{tab:task_emb_for_selection}, we show the effectiveness of task selection methods on
 prompt transfer performance. \bssc{Random} and \bssc{Size} select the source task with the highest task transferability score. \bssc{SEmb-R} and \bssc{Feature} select top-$k$ tasks that exhibit the largest value of the transferability scores. Compared to the no-transfer baseline, these task selection methods show average absolute performance improvements ranging from 0.38\% to 0.91\%. With an increase of the selection pool ($k$=1 to $k$=3), the improvements by \bssc{Semb-R} and \bssc{Feature} further increase to 0.78\% and 1.03\%, respectively. However, the overall transfer gains remain marginal, indicating that the effectiveness of intermediate-task selection is still limited across diverse tasks. %

\subsection{Effect of Task Type and Training Seed}

To dissect the impact of task type and training seed, Table \ref{tab:correlation_task_type} presents the top-3 beneficial intermediate-tasks for \ssc{COPA} and \ssc{CB}. Results for all other tasks are shown in Appendix~\ref{appendix:training_seed_influence}. 

\paragraph{Task type is not a reliable transferability predictor.}
While it is intuitive to assume that similar tasks should transfer well to the downstream task, our results reveal that the top-performing source tasks for a given target task can vary widely in task type.
We find that task types are generally uncorrelated with transfer performances. For example, the most performant source tasks for \ssc{COPA} and \ssc{CB} often come from different task types when various training seeds are used.
Based on three separate runs, the most beneficial source tasks for \ssc{COPA} (QA) are from other task types, such as \ssc{CxC} (semantic similarity) and \ssc{QQP} (paraphrase detection). Similarly, many of the beneficial tasks for \ssc{CB} (NLI) originate from non-NLI tasks.

\paragraph{Random seed significantly impacts the transfer performance.}
For \ssc{COPA}, using different training seeds leads to 7.69\% to 26.78\% relative performance improvements. Similarly, the relative improvements for \ssc{CB} range from 4.11\% to 7.60\%. This emphasizes the crucial role of seed choice in determining transfer performance.
We observe similar variations across seeds in other downstream tasks as well, such as \ssc{CoLA}, \ssc{WiC}, and \ssc{RTE}. This can be attributed to the instability in fine-tuning introduced by different random seeds during prompt transfer~\cite{mosbach2021on,chen-etal-2022-revisiting}, which can largely affect the robustness of intermediate-task selection.

\section{Revisiting the Construction of Task Embeddings}

\label{sec:analysis}

\begin{figure*}[t]
    \centering
    \begin{subfigure}[b]{.46\textwidth}
    \centering
    \includegraphics[width=\textwidth]{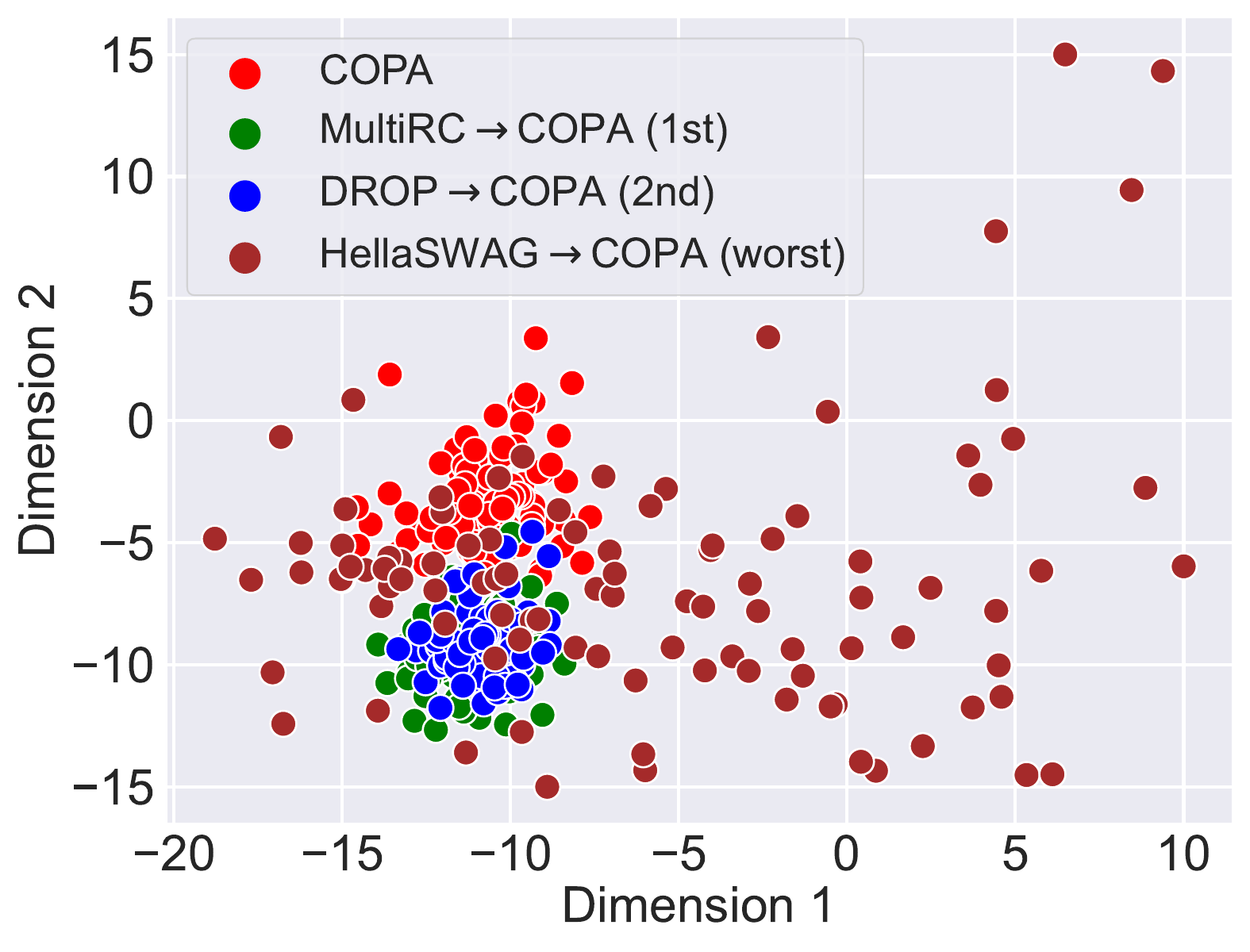}
        \caption{COPA}\label{mycopa}
    \end{subfigure}
    \begin{subfigure}[b]{.46\textwidth}
    \centering 
        \includegraphics[width=\textwidth]{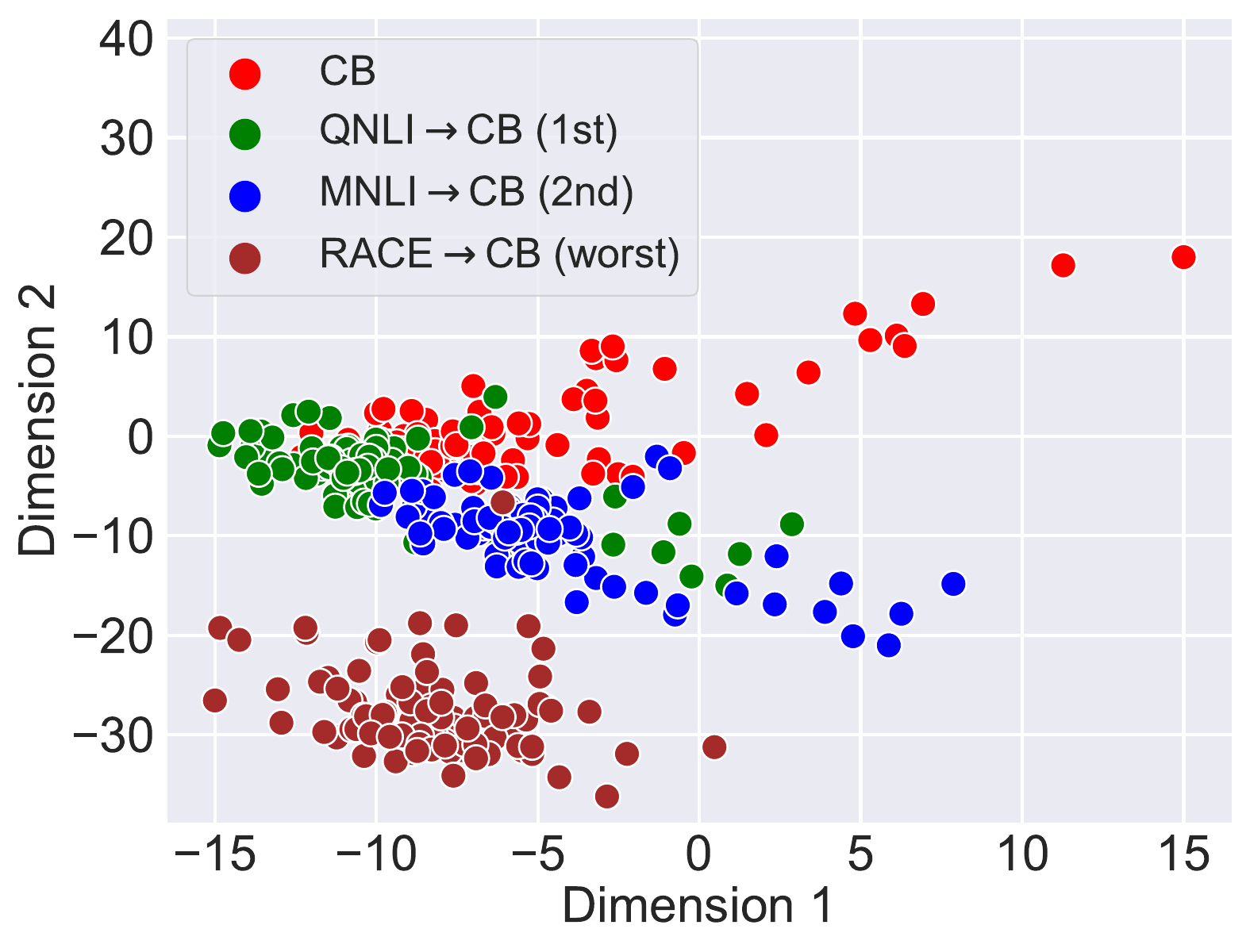}
        \caption{CB}
    \end{subfigure}%
    \caption{
    Projecting prompt tokens of the best, 2nd-best, and worst-performing intermediate-tasks for (\hyperref[mycopa]{a}) COPA and (\hyperref[mycopa]{b}) CB using t-SNE. We observe that prompt tokens from beneficial tasks are distributed more closely to the tokens of no-transfer prompt tuning.}
    \label{fig:token_space}
    \vspace{-3mm}
\end{figure*}

Despite task embeddings from fine-tuned weights demonstrating superior performance in task prediction compared to other selection methods, the effectiveness of various task embedding constructions remains underexplored. 
In this section, we investigate different construction methods of task embeddings. In addition to \bssc{Feature}, we explore two more types of task embeddings as follows.

\subsection{Construction Methods}
\label{learning_curve}

\paragraph{Token-wise mean (\bssc{Unigram})} 
In \bssc{Feature}, we compute the mean of token representations to obtain a task embedding in $\mathbb{R}^{{d}}$. To explore an alternative approach, we compute the task embeddings from another axis, resulting in a task embedding in $\mathbb{R}^{{N}}$. Specifically, the task embedding for a task $t$ denotes as  $h_{t} =\frac{1}{d}[\sum_d{e_1}, \sum_d{e_2}, ..., \sum_d{e_N} ]$. The similarity between tasks $t^1$ and $t^2$ is defined as: $ {\rm sim}(t^{1}, t^{2}) = \cos(h_{t^1}, h_{t^2})$. We refer to this method as \bssc{Unigram} to emphasize that task-specific information is aggregated from the token-wise dimension.

\paragraph{Maximum inner product search (\bssc{Max})}
We propose a novel task embedding method, referred to as \bssc{Max}, based on the maximum token-to-token similarity scores. Given the source task $t^1$ and the target task $t^2$, for each prompt token in $t^2$, we obtain the highest token representation similarity score across all tokens in $t^1$. %
The task similarity is then defined as the mean of these maximum similarity scores:
\begin{align}
    {\rm sim}(t^1, t^2)=\frac{1}{N}\sum_{j=1}^N{\max_i \cos(e_i^1, e_j^2)}
\end{align}

\begin{figure}[t]
\centering
\includegraphics[width=0.48\textwidth]{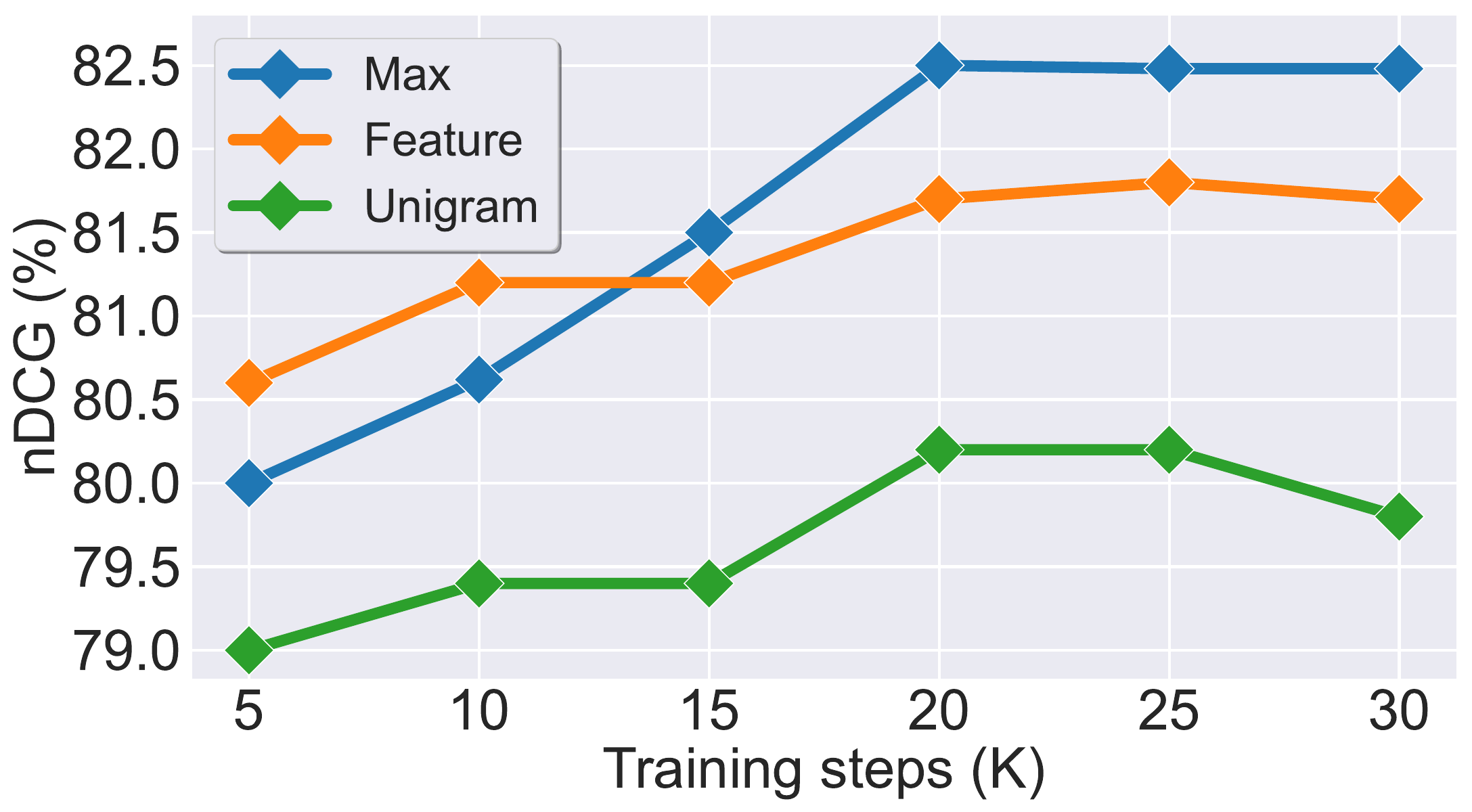}
\vspace{-6mm}
\caption{Task prediction performances (average nDCG scores) of three types of task embeddings.}
\label{fig:transfer_gain}
\vspace{-5mm}
\end{figure}

\subsection{Results and Analysis}
\paragraph{\bssc{Max} achieves the highest task transferability prediction.}
Figure \ref{fig:transfer_gain} presents three types of task embeddings, each derived from prompt checkpoints trained for different numbers of steps.
All three methods show improved performance with longer training steps, suggesting that longer training improves task transferability predictions.
Notably, \bssc{MAX} achieves the highest nDCG score of 82.5\% at the 20K step, indicating that token-wise similarity captures richer task information than \bssc{Feature} and \bssc{Unigram}, leading to more accurate task predictions.

\paragraph{Prompt tokens from beneficial tasks are distributed closer to the target prompt tokens.}

To better understand the prompt token distribution and different levels of transfer performance, we project prompt tokens of the best, 2nd-best, and worst-performing intermediate-tasks onto low-dimensional spaces using t-SNE~\cite{JMLR:v9:vandermaaten08a}.
Figure \ref{fig:token_space} illustrates that the prompt tokens from no-transfer prompt tuning (\textcolor{red}{red}), are close to the tokens from their beneficial intermediate-tasks (\textcolor{mygreen}{green}, \textcolor{darkblue}{blue}).
Furthermore, we observe a considerable overlap in these beneficial source tasks, such as \ssc{MultiRC} and \ssc{DROP}, for downstream task \ssc{COPA}.
This suggests that beneficial tasks tend to be distributed closer to the target prompt tokens and share similar characteristics in low dimensions.
For \ssc{COPA} and \ssc{CB}, the worst-performing intermediate-task (\textcolor{mybrown}{brown}) deviates from the no-transfer prompt tokens. 
Future research can further explore a clearer correlation between intermediate-task token distribution and transfer performance.

\section{Conclusion}

In this work, we conduct a systematic study on intermediate-task selection across a wide range of tasks. 
Our results show that task embeddings based on fine-tuned weights outperform random selection, data size, and text embeddings with improvements of +3.96\%, +3.85\%, and +2.59\% in nDCG scores, underscoring the importance of a task-specific approach.
Nevertheless, we find that task embeddings do not excel in all scenarios, particularly in multiple choice and QA tasks. By revisiting the task embedding construction, we propose a novel method based on pairwise token similarity, which achieves the highest performance of 82.5\% in task transferability prediction, suggesting that token-wise similarity is better predictive in task transferability prediction.\looseness-1

\section*{Limitation}
Despite our proposed method being effective in many scenarios, we observe that it falls short in predicting task transferability for tasks requiring reasoning abilities, which needs to be further explored.
We also face a challenge in precisely evaluating how the parameter configurations of prompt tuning impact transfer performance, as prompt tuning is highly sensitive to hyperparameter selection. Moreover, our evaluation of task selection is limited to one specific model architecture and focused on prompt tuning.
Evaluating different model architectures, model scales, and fine-tuning methods would provide a more comprehensive understanding of the robustness of intermediate-task selection.\looseness-1

\section*{Acknowledgements}
We would like to thank anonymous reviewers for their constructive feedback. This work was funded by the Deutsche Forschungsgemeinschaft (DFG, German Research Foundation) – project-id 232722074 – SFB 1102.

\nocite{pmlr-v97-houlsby19a,pfeiffer-etal-2021-adapterfusion,Zamir_2018_CVPR,kim-etal-2023-taskweb,lee2024instruction,kim-etal-2023-taskweb}

\bibliography{custom,transfer}

\clearpage
\newpage
\appendix

\begin{table*}[ht]
\centering
\begin{adjustbox}{width={\textwidth},totalheight={\textheight},keepaspectratio}
\begin{tabular}{lllllll}
\toprule
\textbf{Name}	&	\textbf{Task} &	\textbf{Task category}	& \textbf{Domain} &	\textbf{|Train|} & \textbf{|Dev|} & \textbf{Metric} \\	
\midrule
\midrule
\emph{13 source tasks}	\\	
\hspace{2mm} \ssc{MNLI} &	\text{NLI}	&	\text{Classification}	& \text{Misc.} &	\text{393K}	& \text{9.8K} & \text{Acc.} \\	
\hspace{2mm} \ssc{QQP} &	\text{Paraphrase detection} &	\text{Classification} & \text{Social QA} &	\text{364K}	& \text{40.4K} & \text{F1/Acc.} \\	
\hspace{2mm} \ssc{QNLI} &	\text{NLI} &	\text{Classification}	& \text{Wikipedia} & \text{105K} & \text{5.4K} & \text{Acc.} \\	

\hspace{2mm} \ssc{ReCoRD} &	\text{QA} &	\text{Multiple Choice}	& \text{News articles} &  \text{101K} & \text{10K} & \text{F1/EM} \\	
\hspace{2mm} \ssc{CxC} &	\text{Semantic similarity}	&	\text{Classification}	& \text{Misc.} & \text{88K} & \text{1K} & \text{Acc.} \\	
\hspace{2mm} \ssc{SQuAD} &	\text{QA}	&	\text{QA}	& \text{Wikipedia, crowd.} & \text{88K} & \text{10.6K} & \text{F1/EM}	\\	
 
\hspace{2mm} \ssc{DROP} &	\text{QA}	&	\text{QA}	& \text{Wikipedia, crowd.} & 	\text{77K} & \text{9.5K} & \text{F1/EM} \\	
\hspace{2mm} \ssc{SST-2} &	\text{Sentiment analysis} &	\text{Classification} & \text{Movie reviews} &	\text{67K} & \text{872} & \text{Acc.} \\	
\hspace{2mm} \ssc{WinoGrande} &	\text{Commonsense reasoning}	&	\text{Multiple Choice}	& \text{Crowdsourced} &	\text{40K} & \text{1.2K} & \text{Acc.} \\	

\hspace{2mm} \ssc{HellaSWAG} &	\text{Commonsense reasoning} &	\text{Multiple Choice} & \text{Misc.} & \text{40K} & \text{10K} & \text{Acc.} \\	
\hspace{2mm} \ssc{MultiRC} &	\text{QA}	&	\text{Classification} & \text{Misc.} & \text{27K} & \text{4.8K} & \text{F1\textsubscript{$\alpha$}/EM} \\	
\hspace{2mm} \ssc{CosmosQA} &	\text{Commonsense reasoning} &	\text{Multiple Choice} & \text{Crowdsourced} & \text{25K} & \text{2.9K} & \text{Acc.} \\	

\hspace{2mm} \ssc{RACE} &	\text{QA}	&	\text{Multiple Choice}	& \text{English exams} & \text{25K} & \text{4.8K} & \text{Acc.} \\	
\midrule
\midrule
\emph{10 target tasks}	\\
\hspace{2mm} \ssc{BoolQ}	& \text{QA} & \text{Classification} & \text{Wikipedia, web queries} & \text{9K} & \text{3.2K} & \text{Acc.} \\	
\hspace{2mm} \ssc{CoLA} & \text{Grammatical acceptability} & \text{Classification} & \text{Books, journals} & \text{9K} & \text{1K} & \text{Matthews cor.} \\
\hspace{2mm} \ssc{STS-B} & \text{Semantic similarity} & \text{Classification} & \text{Misc.} & \text{6K} & \text{1.5K} & \text{Pear./spear.} \\

\hspace{2mm} \ssc{WiC} & \text{Word sense disambiguation} & \text{Classification} & \text{Misc.} & \text{5K} & \text{638} & \text{Acc.} \\
\hspace{2mm} \ssc{CR} & \text{Sentiment analysis} & \text{Classification} & \text{Custom review} & \text{4K} & \text{753} & \text{Acc.} \\
\hspace{2mm} \ssc{MRPC} & \text{Paraphrase detection} & \text{Classification} & \text{News} & \text{4K} & \text{408} & \text{F1/Acc.} \\	

\hspace{2mm} \ssc{RTE} &	\text{NLI}	&	\text{Classification}	& \text{Wikipedia, news} & \text{2K} & \text{277} & \text{Acc.} \\	
\hspace{2mm} \ssc{WSC} & \text{Coreference resolution}& \text{Classification} & \text{Fiction books} & \text{554} & \text{104} & \text{Acc.} \\
\hspace{2mm} \ssc{COPA} & \text{QA} & \text{Multple Choice} & \text{Blog, encyclopedia} & \text{400} & \text{100} & \text{Acc.} \\

\hspace{2mm} \ssc{CB} & \text{NLI} & \text{Classification} & \text{Misc.} & \text{250} & \text{56} & \text{F1/Acc.} \\
\hline
\end{tabular}
\end{adjustbox}\centering{\caption{\label{tab:dataset_all_with_domain_metric}Statistics of source and target tasks. We categorize task types into three types: classification, QA, and multiple choice. We distinguish multiple choice tasks from QA tasks based on whether options are provided in the input.
}}
\end{table*}

\section{More Details to Datasets and Evaluation Metrics}\label{sec:appendix}

\subsection{Datasets}\label{app:detailed_dataset}

We select the datasets drawn from different NLP benchmarks and families of tasks, including natural language inference (NLI), paraphrase detection, semantic similarity, sentiment analysis, question answering (QA), commonsense reasoning, and grammatical acceptability.
In total, we consider 13 source and 10 target tasks.
The distinguishing between high-resource and low-resource tasks follows conventional notions respect with to the training split size.
Table~\ref{tab:dataset_all_with_domain_metric} summarizes the statistics of 23 tasks and the evaluation metrics. 
All data was sourced from HuggingFace Datasets ~\cite{lhoest-etal-2021-datasets}.

\subsection{Evaluation Metircs}
\label{app:metrics}

\paragraph{nDCG}
This metric is built on the concept of Discounted Cumulative Gain (DCG), a measure of the relevance score for a list of items, each discounted by its position in the ranking.

\begin{equation}
    DCG(R) = \sum_{i=1}^{p} \frac{ 2^{rel_{i}}- 1}{\log_{2} (i+1)} 
\end{equation}

where $R$ represents the ranking of source tasks, where the relevance $rel_{i}$ of the source task with rank $i$ is set to the averaged target performance, i.e., $rel_{i} \in [0, 100]$. The ranking position $\rho$ corresponds to the size of the selection budget.

The nDCG is computed as follows:

\begin{equation}
    nDCG(R_{pred}, R_{true}) = \frac{DCG(R_{pred})}{DCG(R_{true})}
\end{equation}

While nDCG generally considers the overall ranking and the difference between predicted transfer performance and actual performance, realistic applications often prioritize the top-1 transfer performance. In this study, our focus is on metrics that accurately quantify the accuracy of top-1 predictions.

\paragraph{Regret@k}
The Regret@k metric is crucial for evaluating how well the task embeddings retrieve the beneficial task for top-1 prompt transfer performance.
Its formula is as follows:

\begin{equation}
    \scalebox{0.9}{$
    \text{Regret@k} = \frac{\max_{s \in S} \mathbb{E}[T(s,t)] - \max_{\Tilde{s} \in S_{k}} \mathbb{E}[T(\Tilde{s},t)]}{O(S)}
    $}
\end{equation}

Now, let's simplify the equation by understanding each term:
$T(s, t)$ represents the performance achieved on the target task $t$ when knowledge is transferred from the source task $s$. In simpler terms, it measures how effective insights from task $s$ are in improving performance on task $t$.
Moving on to $O(S, t)$, this term signifies the expected performance on the target task $t$ under the optimal selection strategy. It establishes a performance benchmark achievable with the most advantageous source task selection.
Finally, consider $M_k(S, t)$, which takes into account the highest performance observed on task $t$ among the $k$ top-ranked source tasks. This aspect evaluates the potential of the selected set of source tasks in contributing to superior performance on the target task $t$.

\section{Transfer Gains with Varying Training Data Sizes}\label{appendix:transfer_gain_with_data_size}
\begin{figure}[ht!]
    \centering
    \includegraphics[scale=0.5]{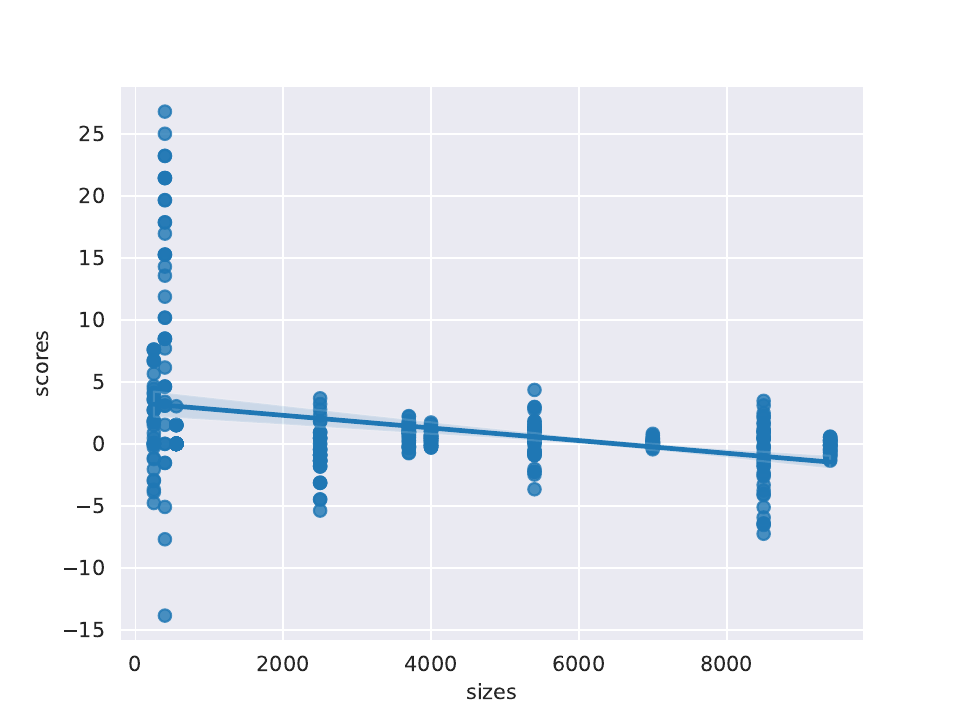}
    \caption{
    Transfer gains with soft prompt transfer.
    The dot on y-axis indicates the number of improved transfer performances compared to prompt tuning, while the x-axis enumerates the training set sizes on 10 downstream tasks.}\label{fig:transfer_gain_vs_sizes}
\end{figure}

We further explore how the training data size influences the relative performance.
Figure \ref{fig:transfer_gain_vs_sizes} illustrates the correlation between the training split size and the level of transfer gains and losses. 
The plot shows 39 runs for each target task.
Remarkably, tasks with extremely low resources (fewer than 1K training samples) exhibit a broad range of transfer gains and losses.
Specifically, Tasks like \ssc{COPA} and \ssc{CB} with minimal training samples (400 and 250, respectively) show transfer gains varying from +25\% to -15\% in relative performance.

On the other hand, tasks with smaller variance in transfer gains, such as \ssc{WSC} and \ssc{RTE}, tend to have fewer instances of positive transfer.
This is influenced by a substantial number of runs achieving similar performance to baselines, leading to fewer positive transfers.
Additionally, our prompt tuning settings, optimized for near-optimal performance, result in less pronounced benefits from prompt training.

The mean slope emphasizes trends, highlighting a strong correlation between the number of positive gains and the training sample sizes across most downstream tasks. Notably, the extent of performance improvement is more significant for tasks with smaller training sample sizes. However, despite high variance in relative performance, transfer gains tend to converge to zero when the dataset size reaches around 5K.

Prompt transfer's success is intricately tied to the data size of downstream tasks. Smaller training examples are more likely to exhibit positive transfer. While prompt transfer brings benefits, the presence of negative transfer underscores associated risks.

\begin{table*}[ht]
\centering
\begin{adjustbox}{max width=\textwidth}
\begin{tabular}{ l l l l l l l l l l l}
\toprule
& \bssc{BoolQ} & \bssc{CoLA} & \bssc{STS-B} & \bssc{WiC} & \bssc{CR} & \bssc{MRPC} & \bssc{RTE}  & \bssc{WSC} & \bssc{COPA} & \bssc{CB} \\
\bssc{Prompt-Abstract} & 
{73.0}$_{1.2}$ &
{52.9$_{1.2}$} & {88.1}$_{0.6}$ &
{63.6}$_{1.6}$ &
{93.5}$_{0.2}$ & {86.1}$_{0.7}$ & {68.7}$_{1.2}$ &  {71.5}$_{1.7}$ & {56.7}$_{1.7}$ & {92.7}$_{1.9}$ \\

\bssc{Prompt-Text} & 
\textcolor{myorange}{78.69}$_{0.18}$ & \textcolor{myorange}{62.47}$_{1.51}$ & \textcolor{myorange}{90.14}$_{0.20}$ & \textcolor{myorange}{69.07}$_{0.45}$ & \textcolor{myorange}{92.96}$_{0.29}$ & \textcolor{myorange}{89.95}$_{0.52}$ & \textcolor{myorange}{79.66}$_{0.74}$ & \textcolor{myorange}{63.46}$_{0.00}$ & \textcolor{myorange}{60.0}$_{3.74}$ & \textcolor{myorange}{85.64}$_{2.21}$ \\
\midrule
\bssc{MNLI} & \textcolor{black}{78.36}$_{0.20}$ & \textcolor{black}{61.55}$_{0.70}$ & \hlc[lightgreen]{90.22$_{0.16}$} & \textcolor{black}{69.07}$_{0.32}$ & \hlc[lightgreen]{{93.18}$_{0.31}$} & \hlc[lightgreen]{{90.93}$_{0.16}$} & \textcolor{black}{78.45}$_{0.45}$ & \textcolor{black}{63.46}$_{0.00}$ & \hlc[lightgreen]{{63.00}$_{5.09}$} & \hlc[lightgreen]{{87.62}$_{2.79}$} \\
\bssc{QQP} & \textcolor{black}{78.66}$_{0.09}$ & \textcolor{black}{61.68}$_{0.86}$ & \hlc[lightgreen]{{90.29}$_{0.15}$} & \textcolor{black}{68.44}$_{0.29}$ & \textcolor{black}{92.96}$_{0.21}$ & \hlc[lightgreen]{{90.69}$_{0.15}$} & \hlc[lightgreen]{{80.14}$_{0.88}$} & \textbf{\hlc[darkgreen]{{64.42}$_{0.78}$}} & \hlc[lightgreen]{{67.33}$_{2.86}$} & \textcolor{black}{84.72}$_{1.02}$ \\
\bssc{QNLI} & 
\textbf{\hlc[darkgreen]{{78.80$_{0.15}$}}} & \textcolor{black}{61.97}$_{0.79}$ & \textcolor{black}{90.04}$_{0.13}$ & \textcolor{black}{68.39}$_{0.14}$ & \textbf{\hlc[darkgreen]{{93.80$_{0.16}$}}}& \hlc[lightgreen]{{90.49}$_{0.37}$} & \textcolor{black}{77.61}$_{0.77}$ & \textcolor{black}{63.46}$_{0.00}$ & \hlc[lightgreen]{{61.33}$_{3.77}$} & \hlc[lightgreen]{{88.67}$_{1.50}$} \\

\bssc{ReCoRD} & \textcolor{black}{78.27}$_{0.18}$ & \textcolor{black}{60.31}$_{0.23}$ & \hlc[lightgreen]{90.36$_{0.10}$} & \hlc[lightgreen]{69.64$_{0.63}$} & \hlc[lightgreen]{93.05$_{0.06}$} & \hlc[lightgreen]{90.65$_{0.47}$} & \textcolor{black}{79.18}$_{0.61}$ & \hlc[lightgreen]{63.78$_{0.45}$} & \hlc[lightgreen]{67.67$_{1.70}$} & \textbf{\hlc[darkgreen]{89.29$_{0.73}$}} \\
\bssc{CxC} & \textcolor{black}{78.71$_{0.25}$} & \hlc[lightgreen]{{62.49$_{0.82}$}} & \textcolor{black}{90.12$_{0.11}$} & \hlc[lightgreen]{69.59$_{1.22}$} & \hlc[lightgreen]{93.45$_{0.35}$} & \hlc[lightgreen]{90.62$_{0.21}$} & \textcolor{black}{79.30$_{1.12}$} & \textcolor{black}{63.46$_{0.00}$} & \hlc[lightgreen]{68.00$_{0.82}$} & \hlc[lightgreen]{86.60$_{2.06}$} \\
\bssc{SQuAD} & \textcolor{black}{\hlc[lightgreen]{78.80$_{0.28}$}} & \textcolor{black}{61.43}$_{1.43}$ & \textcolor{black}{\hlc[lightgreen]{90.17$_{0.08}$}} & \textcolor{black}{\hlc[lightgreen]{69.49$_{0.77}$}} & \textcolor{black}{\hlc[lightgreen]{93.63$_{0.38}$}} & \textcolor{black}{\hlc[lightgreen]{90.41$_{0.28}$}} & \textcolor{black}{77.74$_{1.33}$} & \textcolor{black}{\hlc[lightgreen]{63.78$_{0.45}$}} & \textcolor{black}{\hlc[lightgreen]{65.67$_{1.25}$}} & \textcolor{black}{\hlc[lightgreen]{87.15$_{3.44}$}} \\

\bssc{DROP} & \textcolor{black}{78.37$_{0.46}$} & \textcolor{black}{61.01$_{0.17}$} & \textcolor{black}{\hlc[lightgreen]{90.23$_{0.10}$}} & \textcolor{black}{\hlc[lightgreen]{69.12$_{0.80}$}} & \textcolor{black}{\hlc[lightgreen]{93.71$_{0.23}$}} & \textbf{\hlc[darkgreen]{91.22$_{0.47}$}} & \textcolor{black}{\hlc[lightgreen]{80.39$_{0.45}$}} & \textcolor{black}{63.46$_{0.00}$} & \textcolor{black}{\hlc[lightgreen]{67.00$_{2.16}$}} & \textcolor{black}{\hlc[lightgreen]{86.37$_{2.37}$}} \\

\bssc{SST-2} & \textcolor{black}{78.56$_{0.33}$} & \textcolor{black}{61.36$_{0.73}$} & \textcolor{black}{89.91$_{0.14}$} & \textcolor{black}{\hlc[lightgreen]{69.64$_{0.60}$}} & \textcolor{black}{\hlc[lightgreen]{93.54$_{0.41}$}} & \textcolor{black}{\hlc[lightgreen]{90.35$_{0.05}$}} & \textcolor{black}{78.46$_{1.12}$} & \textcolor{black}{\hlc[lightgreen]{63.78$_{0.45}$}} & \textcolor{black}{\hlc[lightgreen]{61.67$_{1.70}$}} & \textcolor{black}{\hlc[lightgreen]{86.93$_{0.39}$}} \\

\bssc{WinoGrande} & \textcolor{black}{78.42$_{0.13}$} & \textcolor{black}{\hlc[lightgreen]{62.72$_{1.02}$}} & \textcolor{black}{\hlc[lightgreen]{90.19$_{0.11}$}} & \textcolor{black}{\hlc[lightgreen]{69.70$_{1.04}$}} & \textcolor{black}{92.87$_{0.17}$} & \textcolor{black}{\hlc[lightgreen]{90.98$_{0.44}$}} & \textcolor{black}{79.18$_{1.23}$} & \textcolor{black}{63.46$_{0.00}$} & \textcolor{black}{\hlc[lightgreen]{67.67$_{1.25}$}} & \textcolor{black}{\hlc[lightgreen]{87.05$_{2.40}$}} \\

\bssc{HellaSWAG} & \textcolor{black}{78.42$_{0.30}$} & \textbf{\textcolor{black}{\hlc[darkgreen]{63.04$_{1.32}$}}} & \textbf{\textcolor{black}{\hlc[darkgreen]{90.46$_{0.10}$}}} & \textcolor{black}{\hlc[lightgreen]{69.38$_{0.77}$}} & \textcolor{black}{\hlc[lightgreen]{93.23$_{0.11}$}} & \textcolor{black}{\hlc[lightgreen]{90.59$_{0.25}$}} & \textcolor{black}{78.70$_{0.59}$} & \textcolor{black}{\hlc[lightgreen]{63.78$_{0.45}$}} & \textcolor{black}{\hlc[lightgreen]{63.33$_{5.25}$}} & \textcolor{black}{\hlc[lightgreen]{85.75$_{2.05}$}} \\
\bssc{MultiRC} & \textcolor{black}{78.69$_{0.02}$} & \textcolor{black}{62.26$_{0.46}$} & \textcolor{black}{90.13$_{0.15}$} & \textcolor{black}{\hlc[lightgreen]{69.59$_{0.22}$}} & \textcolor{black}{\hlc[lightgreen]{93.14$_{0.27}$}} & \textcolor{black}{\hlc[lightgreen]{90.37$_{0.12}$}} & \textcolor{black}{79.06$_{1.53}$} & \textcolor{black}{\hlc[lightgreen]{63.78$_{0.45}$}} & \textcolor{black}{\hlc[lightgreen]{68.00$_{2.16}$}} & \textcolor{black}{\hlc[lightgreen]{87.63$_{0.31}$}} \\
\bssc{CosmosQA} & \textcolor{black}{78.47$_{0.24}$} & \textcolor{black}{61.40$_{0.52}$} & \textcolor{black}{90.10$_{0.06}$} & \textbf{\textcolor{black}{\hlc[darkgreen]{70.22$_{1.02}$}}} & \textcolor{black}{\hlc[lightgreen]{93.63$_{0.11}$}} & \textcolor{black}{\hlc[lightgreen]{90.96$_{0.20}$}} &
\textbf{\textcolor{black}{\hlc[darkgreen]{80.63$_{1.04}$}}} &
\textcolor{black}{63.46$_{0.00}$} & \textcolor{black}{\hlc[lightgreen]{66.67$_{1.25}$}} & \textcolor{black}{\hlc[lightgreen]{87.46$_{0.38}$}} \\
\bssc{RACE} & \textcolor{black}{78.24$_{0.43}$} & \textcolor{black}{61.05$_{1.42}$} & \textcolor{black}{\hlc[lightgreen]{90.16$_{0.11}$}} & \textcolor{black}{68.70$_{1.93}$} & \textcolor{black}{\hlc[lightgreen]{93.67$_{0.13}$}} & \textcolor{black}{\hlc[lightgreen]{90.67$_{0.33}$}} & \textcolor{black}{\hlc[lightgreen]{80.39$_{0.90}$}} & \textcolor{black}{63.46$_{0.00}$} & \textbf{\textcolor{black}{\hlc[darkgreen]{68.00$_{0.00}$}}} & \textcolor{black}{\hlc[lightgreen]{88.07$_{2.56}$}} \\

\bottomrule
\end{tabular}
\end{adjustbox}
\caption{Results of prompt transfer. Downstream task performances involve soft prompt transfer between intermediate tasks (rows) and target tasks (columns) using the T5 base model. 
The first two rows represent the baseline performances with prompt tuning, without any pre-trained prompt weights.
\ssc{Prompt-Abstract} refers to prompt tuning with the abstract symbol as a class label, and \ssc{Prompt-Text} refers to prompt tuning using the text span.
Subsequent rows provide insights into prompt transfer performances, where the best-performing prompts from each task are transferred to ten different downstream tasks.
All reported scores are mean values obtained from three random restarts.}
\label{transfer_results}
\end{table*}

\section{Prompt Transfer Performance}\label{appendix:emprical_prompt_transfer_performance}
Table \ref{transfer_results} presents the mean performance across three runs on low-resource tasks, utilizing the best-performing soft prompt as the initialization point. 
As seen in previous studies, the prompt transfer results indicate improvements over the no-transfer baselines.

In particular, our most successful transfer results exhibit significant enhancements, surpassing the no-transfer outcomes on tasks such as \ssc{COPA} and \ssc{CB} by considerable margins, with improvements of +$8$\% and $+3.46$\%, respectively.
However, it's noteworthy that the mean performance improvements for other tasks are relatively minor. This can be attributed to the extensive hyperparameter search conducted for the strong baseline (\ssc{Prompt-Text}), contrasting with the suboptimal nature of the weak baseline (\ssc{Prompt-Abstract}). 
This underlines the significance of optimization in the prompt tuning process.

Our exploration of prompt transfer performance sheds light on the nuanced dynamics at play, emphasizing the need for strategic optimization strategies in achieving robust and notable improvements, especially in the context of low-resource tasks.

\begin{table*}[ht]
\centering
\begin{adjustbox}{max width=\textwidth}
\begin{tabular}{p{3.3cm}|lll|lll|lll}
\toprule
\multirow{2}{*}{\textbf{Target}} & \multicolumn{3}{c|}{\textbf{seed 112}} & \multicolumn{3}{c|}{\textbf{28}} & \multicolumn{3}{c}{\textbf{52}} \\
& \textbf{Source} & \textbf{Task Type} & \textbf{Rel. (\%)} & \textbf{Source} & \textbf{Task Type} & \textbf{Rel. (\%)} & \textbf{Source} & \textbf{Task Type} & \textbf{Rel. (\%)} \\
\midrule
\emph{Top-3 transfer} & \multicolumn{3}{c|}{} & \multicolumn{3}{c|}{} & \multicolumn{3}{c}{} \\
\multirow[t]{3}{3cm}{\hspace{1mm} BoolQ (QA)} & DROP* & QA & 0.58 & SQuAD* & QA & 0.31 & CxC & senti. similarity & 0.31 \\
 & SST-2 & sentiment & 0.55 & QQP & paragraph & -0.12 & QNLI & NLI & 0.27 \\
& HellaSWAG	& commonsense & 0.50 & QNLI & NLI & -0.15	& SQuAD* & QA & 0.11
 \\
\midrule
\multirow[t]{3}{3cm}{\hspace{1mm} CoLA (grammatical acceptability)} & WinoGrande & commonsense & 3.47 & WinoGrande & commonsense & 3.10 & HellaSWAG & commonsense & 0.44 \\
 & RACE & QA & 2.47 & CxC & senti. similarity & 2.10 & CxC & senti. similarity & -1.79 \\
 & MultiRC & QA & 2.24 & QQP & paragraph & 1.65 & QNLI & NLI & -2.35 \\
\midrule
\multirow[t]{3}{3cm}{\hspace{1mm} STS-B (sentiment similarity)} & ReCoRD & QA & 0.16 & ReCoRD & QA & 0.18 & HellaSWAG & commonsense & 0.81 \\
 & HellaSWAG & commonsense & 0.08 & HellaSWAG & commonsense & 0.16 & QQP & paragraph & 0.68 \\
 & DROP & QA & 0.07 & WinoGrande & commonsense & 0.08 & MultiRC & QA & 0.52 \\
\midrule
\multirow[t]{3}{3cm}{\hspace{1mm} WiC (word sense disambiguation)} & WinoGrande & commonsense & 2.95 & ReCoRD & QA & 1.35 & CosmosQA & commonsense & 4.35 \\
 & CxC & senti. similarity & 1.81 & SQuAD & QA & 1.13 & CxC & senti. similarity & 2.98 \\
 & CosmosQA & commonsense & 1.59 & SST-2 & sentiment & 0.90 & HellaSWAG & commonsense & 2.75 \\
\midrule
\multirow[t]{3}{3cm}{\hspace{1mm} CR (sentiment)} & SST-2* & sentiment & 0.71 & SQuAD & QA & 1.72 & DROP & QA & 1.00 \\
 & CosmosQA/RACE & commonsense/QA & 0.57 & QNLI & NLI & 1.58 & QNLI/SST-2* & NLI/sentiment & 0.71 \\
 & MNLI/QNLI & NLI/NLI & 0.43 & CxC & senti. similarity & 1.29 & CxC/CosmosQA & senti. similarity/commonsense & 0.57 \\
\midrule
\multirow[t]{3}{3.3cm}{\hspace{1mm} MRPC (paraphrase)} & DROP & QA & 2.24 & WinoGrande & commonsense & 2.19 & DROP & QA & 0.95 \\
 & CosmosQA & commonsense & 1.85 & RACE & QA & 1.68 & MNLI & NLI & 0.48 \\
 & QQP* & paragraph & 1.75 & ReCoRD & QA & 1.66 & CosmosQA & commonsense & 0.27 \\
\midrule
\multirow[t]{3}{3cm}{\hspace{1mm} RTE (NLI)} & MultiRC & QA & 1.81 & RACE & QA & 3.67 & CosmosQA & commonsense & 1.79 \\
 & QQP/RACE & paragraph/QA & 0.45 & QQP & paragraph & 3.21 & CxC/WinoGrande & senti. similarity/commonsense & 0.45 \\
 & DROP & QA & 0.00 & DROP & QA & 2.75 & DROP & QA & 0.00 \\
\midrule
\multirow[t]{3}{3cm}{\hspace{1mm} WSC$^{\dag}$ (coreference resolution)} & QQP/SQuAD/SST-2 & paragraph/QA/sentiment & 1.52 & ReCoRD/MultiRC & QA/QA & 1.52 & QQP & paragraph & 3.03 \\
 & MNLI/QNLI & NLI/NLI & 0.00 & MNLI/QQP/QNLI & NLI/paraphrase/NLP & 0.00 & MNLI/QNLI & NLI/NLI & 0.00 \\
 & - & - & - & - & - & - & - & - & - \\
\bottomrule
\end{tabular}
\end{adjustbox}
\caption{Top-3 prompt transfer on eight downstream target tasks and their task types. The three most significant improvements in prompt transfer across three random seeds, 112, 28, and 52. The relative performance is reported as a percentage (\%) and calculated based on the corresponding no-transfer prompt-tuning. * indicates that the source task type is identical to the task type of the downstream task.}\label{top_3_prompt_transfer_results_on_eight_task}
\end{table*}

\section{More Results on the Effect of Task Type and Training Seed}\label{appendix:training_seed_influence}

Table~\ref{top_3_prompt_transfer_results_on_eight_task} presents the top three prompt transfer results on eight downstream target tasks, along with their respective task types. These results reflect the most significant improvements in prompt transfer across three random seeds. On tasks with limited annotations, such as \ssc{COPA} and \ssc{CB}, different random seeds lead to substantial variance in transfer performance. 
Similarly, tasks like \ssc{CoLA}, \ssc{WIC}, and \ssc{RTE} also exhibit high variance. For \ssc{WSC}$^{\dag}$, we observed that most prompt transfer performances either present identical transfer gain or show no improvement in performance. This phenomenon is likely attributed to the unique task type of \ssc{WSC} compared to other downstream tasks. Specifically, the knowledge of source tasks has limited influence on performing the tasks.

\section{More Results on the Construction of Task Embeddings}
\begin{figure*}[t]
    \centering
    \begin{subfigure}[b]{0.44\linewidth}
        \centering
        \includegraphics[width=\linewidth]{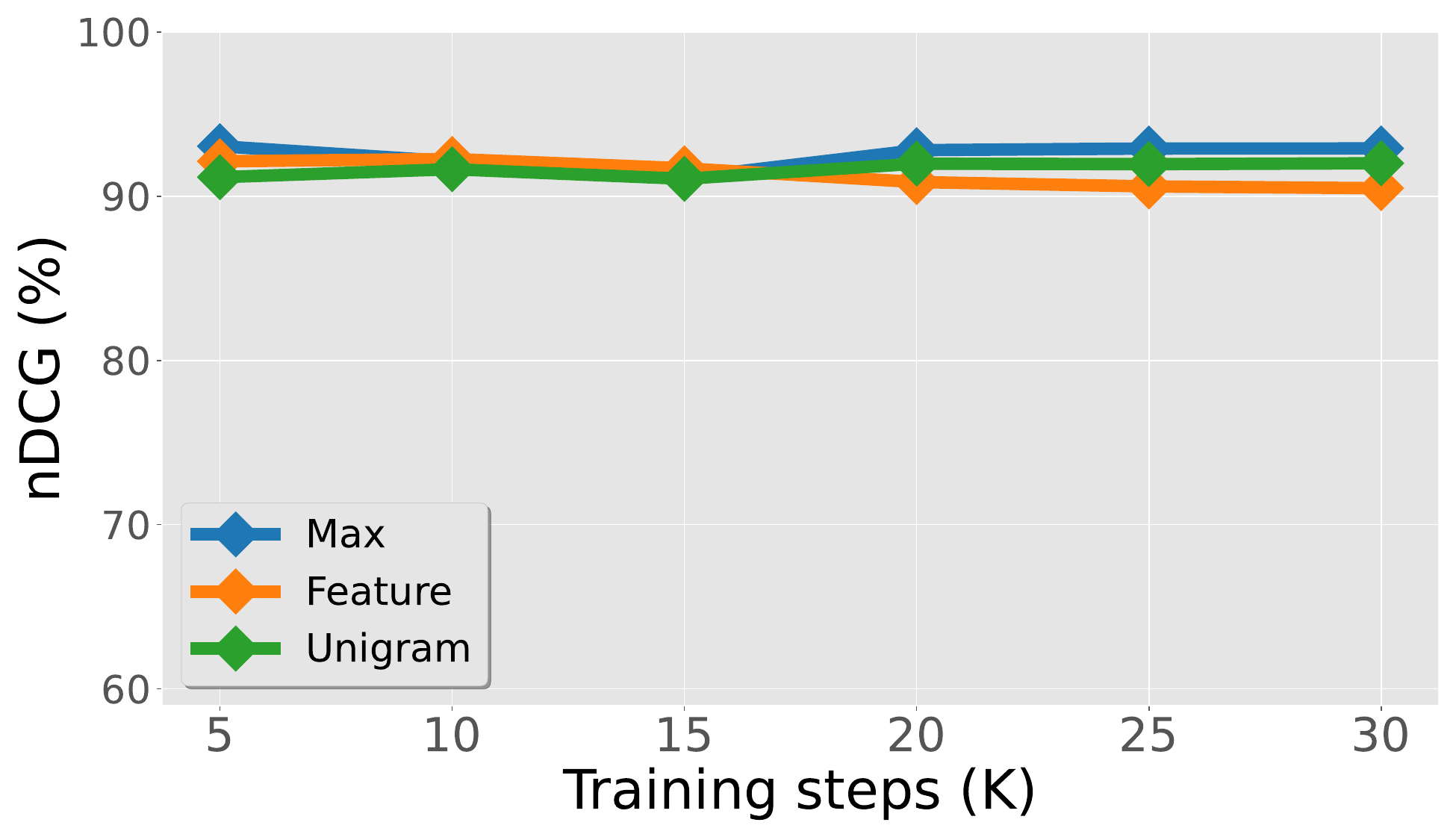}
        \caption{BoolQ}\label{ndcg_boolq}
    \end{subfigure}%
    \begin{subfigure}[b]{0.44\linewidth}
        \centering
        \includegraphics[width=\linewidth]{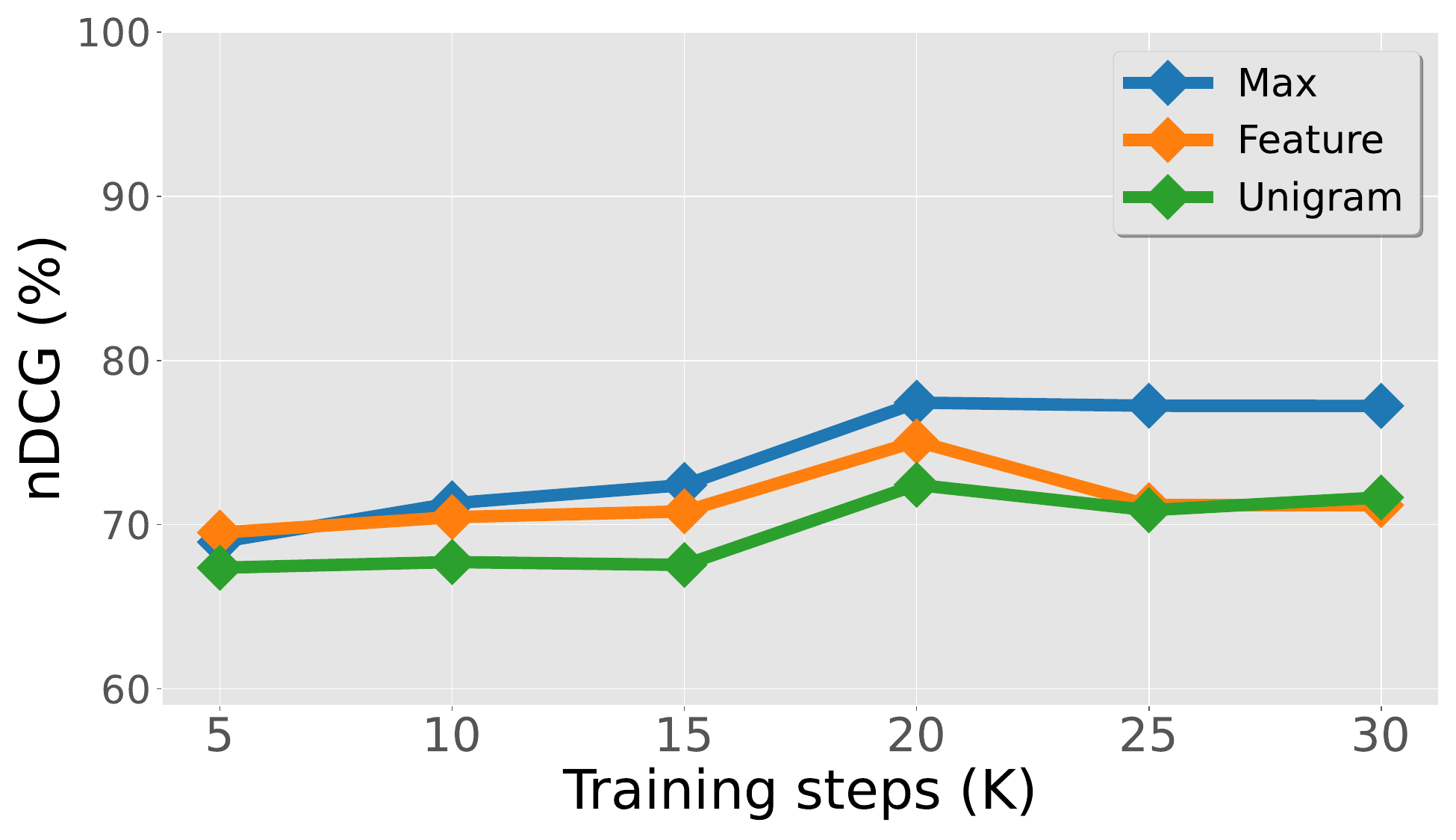}
        \caption{CoLA}\label{ndcg_cola}
    \end{subfigure}%
    \hspace{0.1em}
    \begin{subfigure}[b]{0.44\linewidth}
        \centering
        \includegraphics[width=\linewidth]{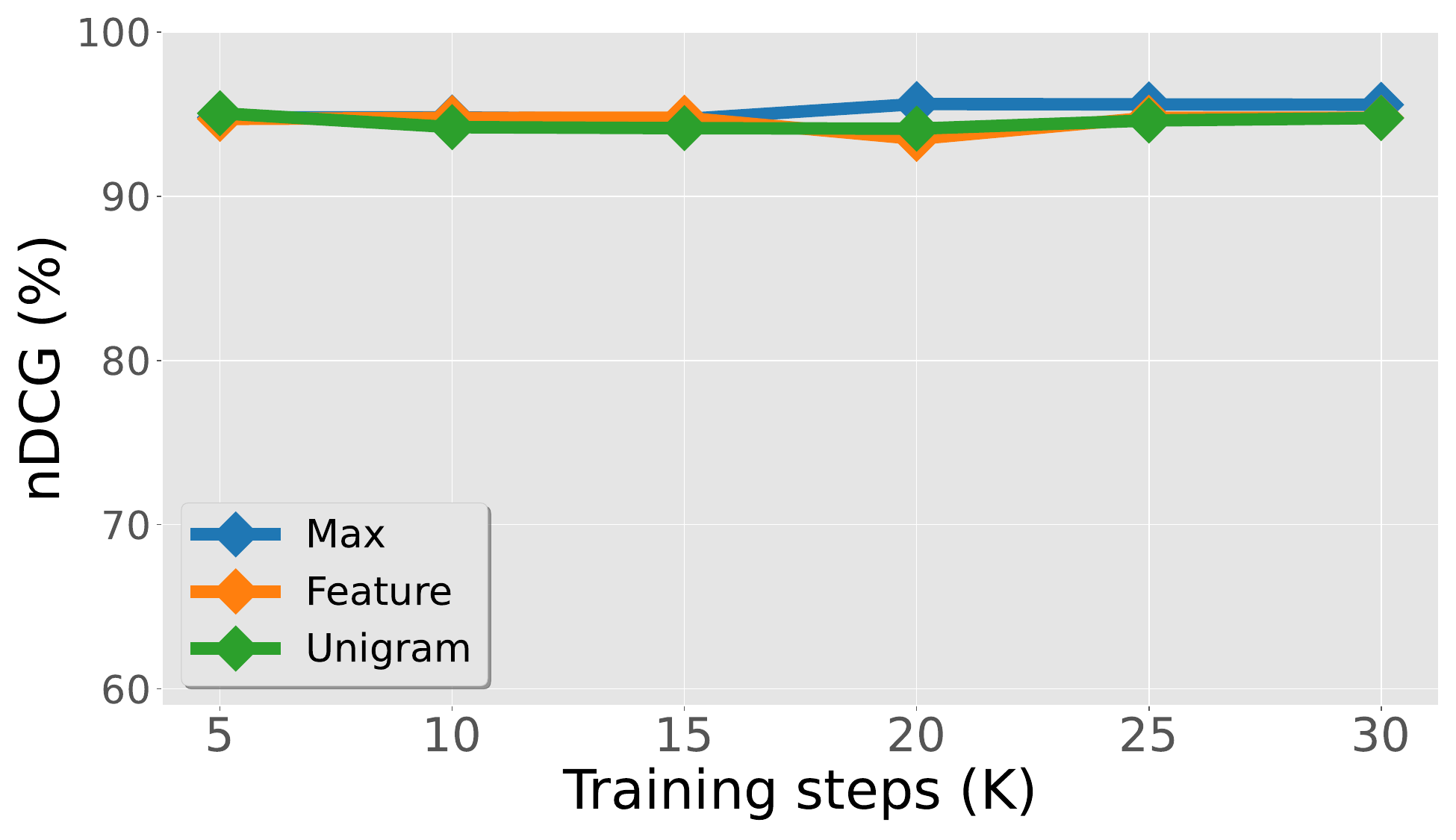}
        \caption{STS-B}\label{ndcg_stsb}
    \end{subfigure}%
    \begin{subfigure}[b]{0.44\linewidth}
        \centering
        \includegraphics[width=\linewidth]{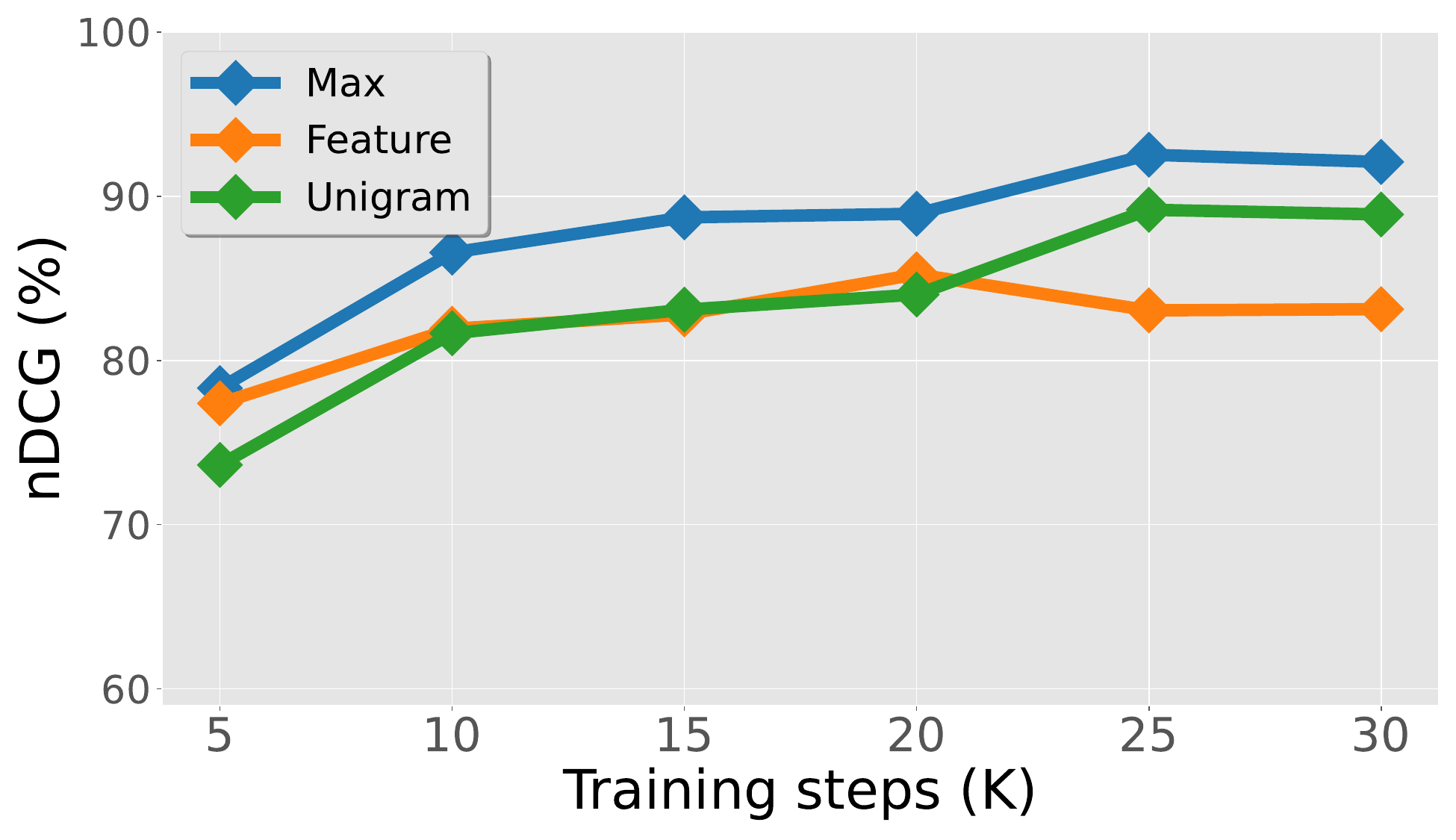}
        \caption{WiC}\label{ndcg_wic}
    \end{subfigure}%
    \hspace{0.1em}
    \begin{subfigure}[b]{0.44\linewidth}
        \centering
        \includegraphics[width=\linewidth]{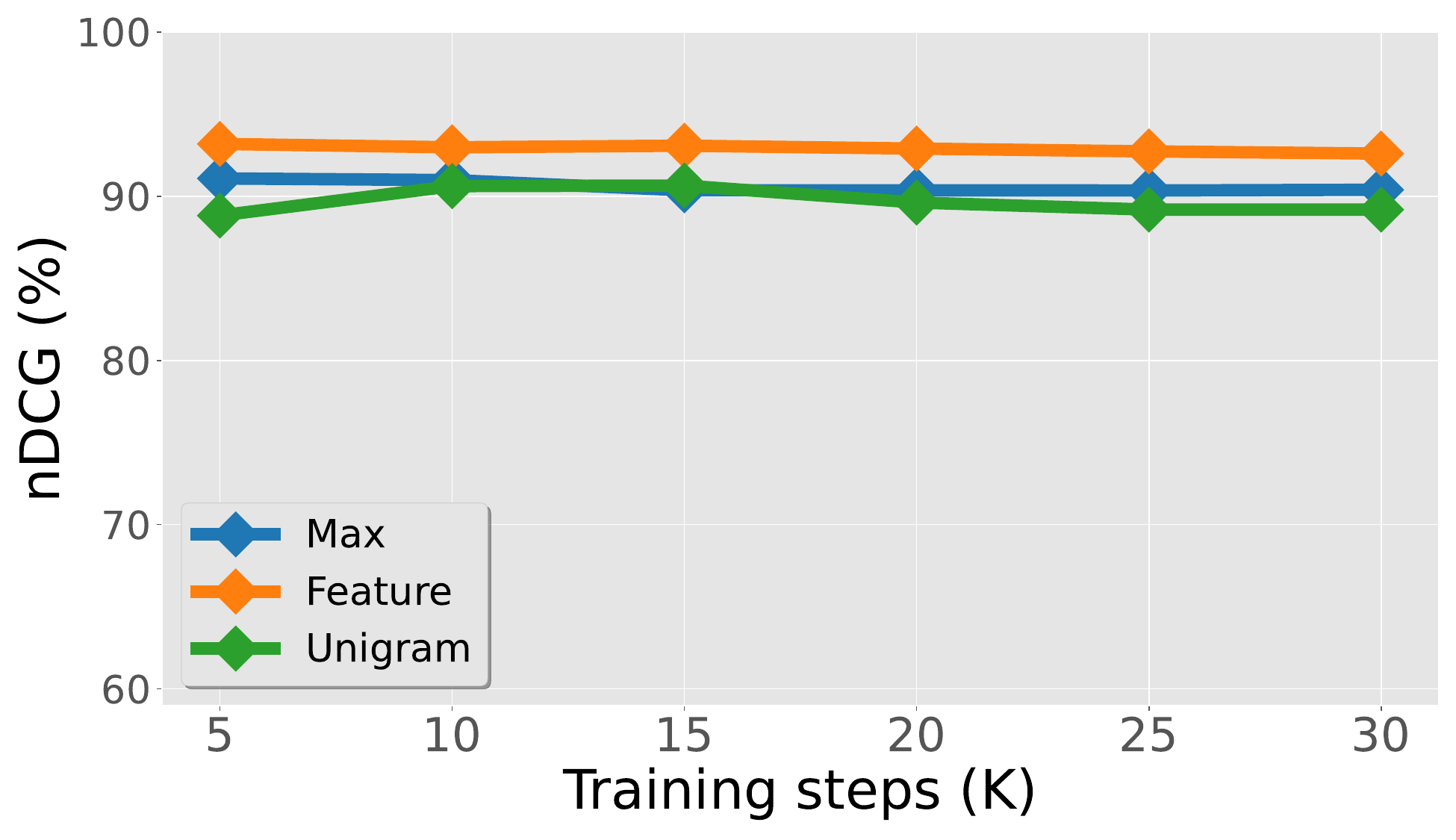}
        \caption{CR}\label{ndcg_cr}
    \end{subfigure}%
    \begin{subfigure}[b]{0.44\linewidth}
        \centering
        \includegraphics[width=\linewidth]{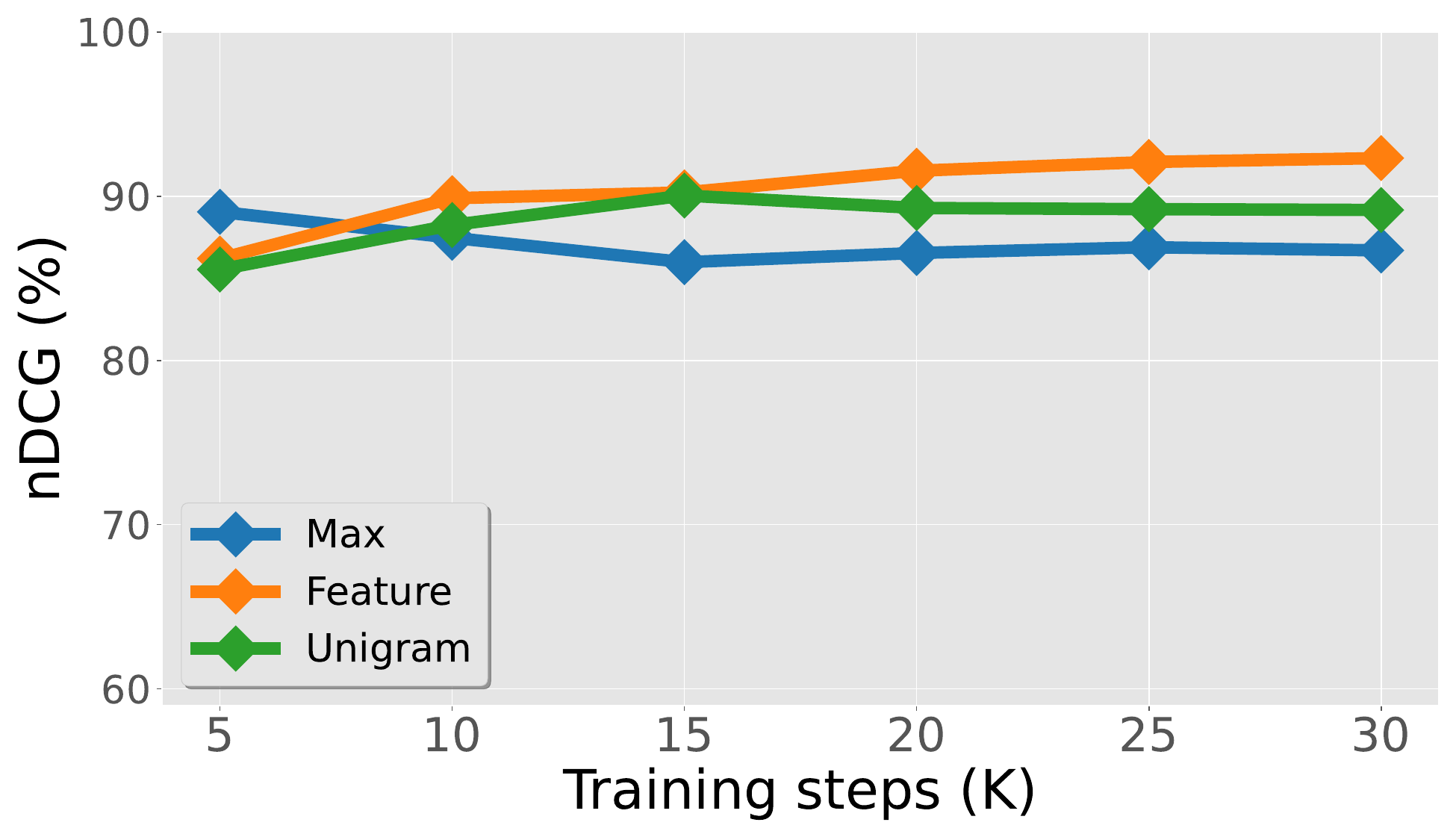}
        \caption{MRPC}\label{ndcg_mrpc}
    \end{subfigure}%
    \hspace{0.1em}
    \begin{subfigure}[b]{0.44\linewidth}
        \centering
        \includegraphics[width=\linewidth]{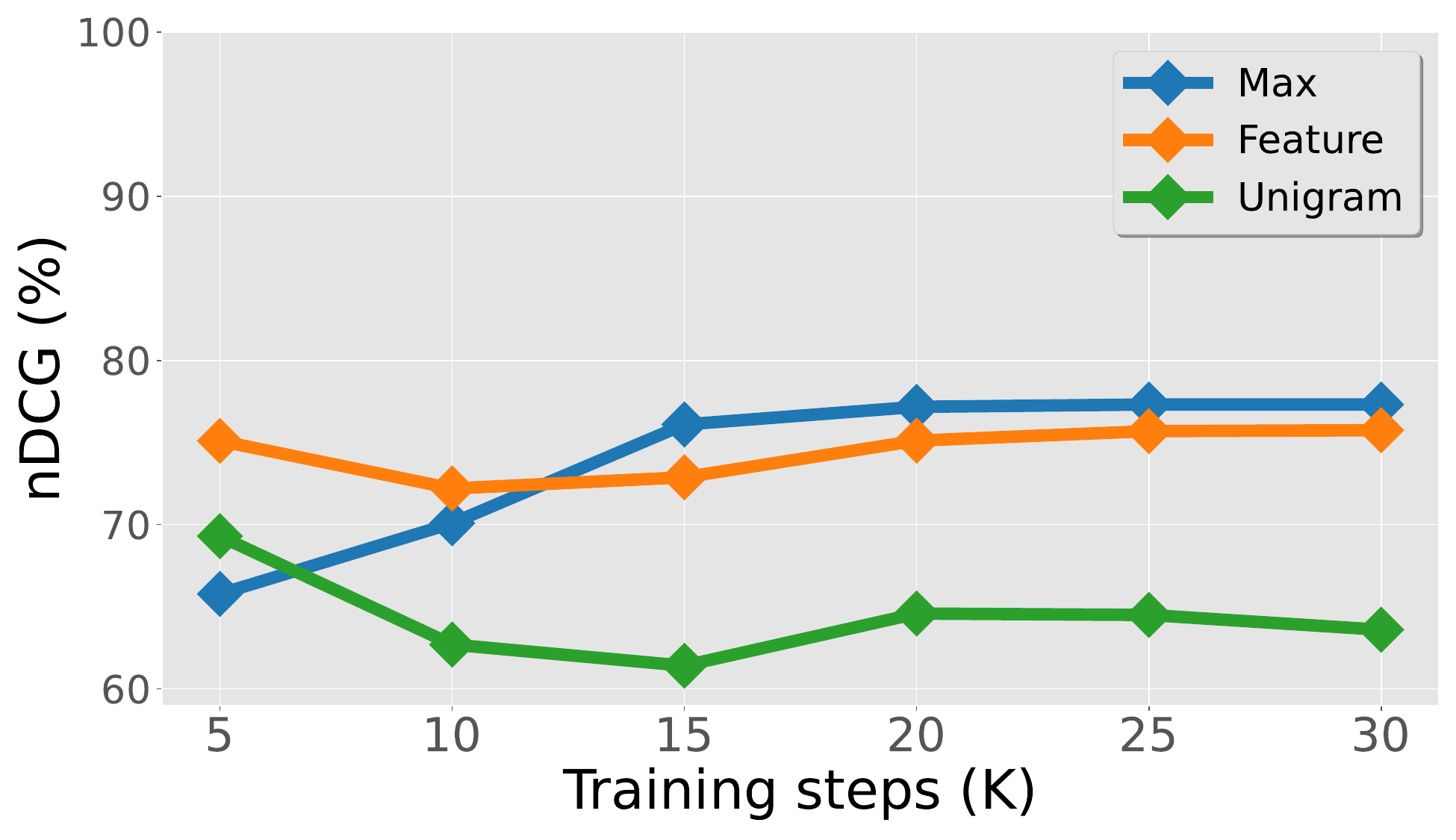}
        \caption{RTE}\label{ndcg_rte}
    \end{subfigure}%
    \begin{subfigure}[b]{0.44\linewidth}
        \centering
        \includegraphics[width=\linewidth]{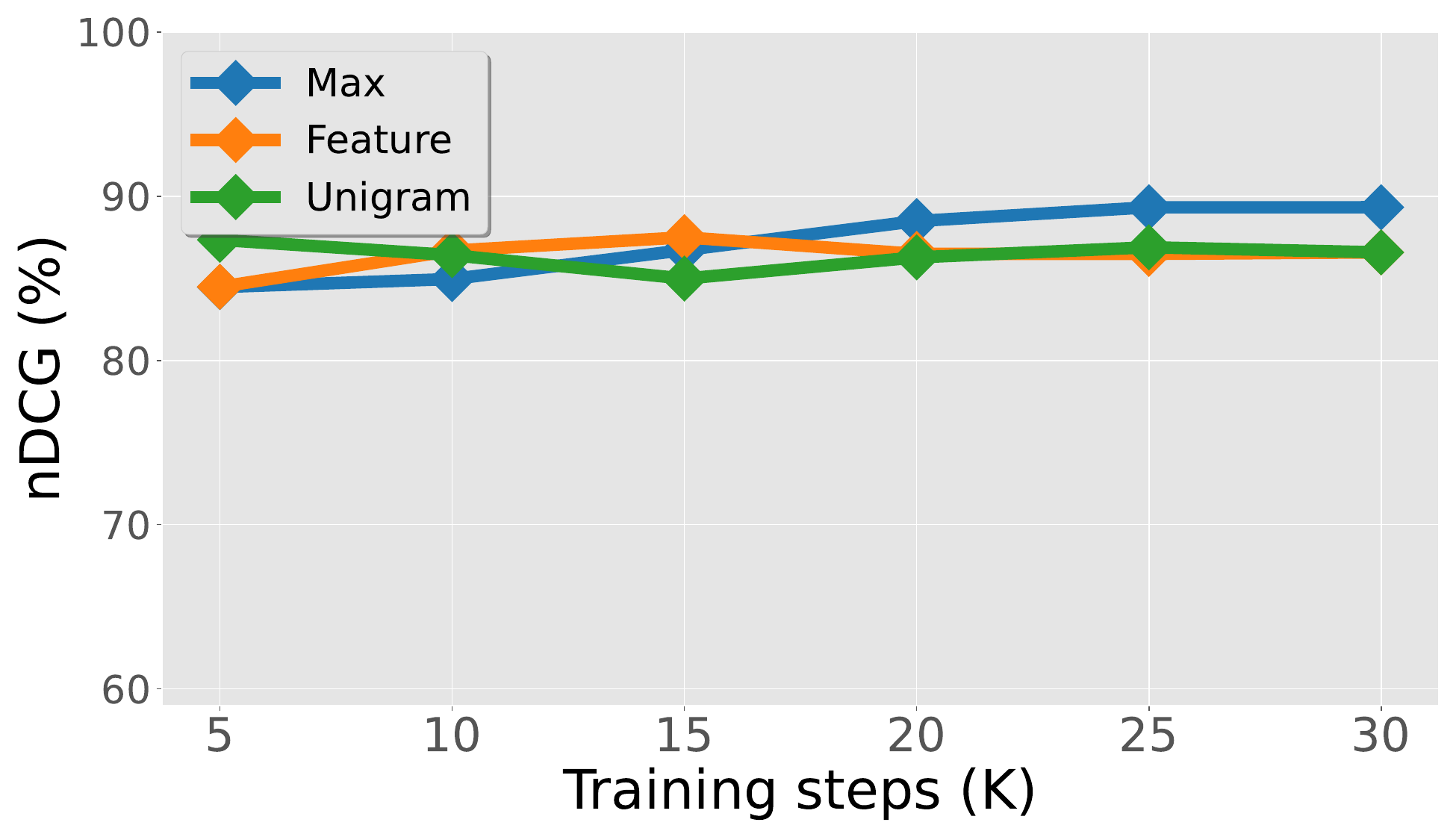}
        \caption{WSC}\label{ndcg_wsc}
    \end{subfigure}%
    \hspace{0.1em}
    \begin{subfigure}[b]{0.44\linewidth}
        \centering
        \includegraphics[width=\linewidth]{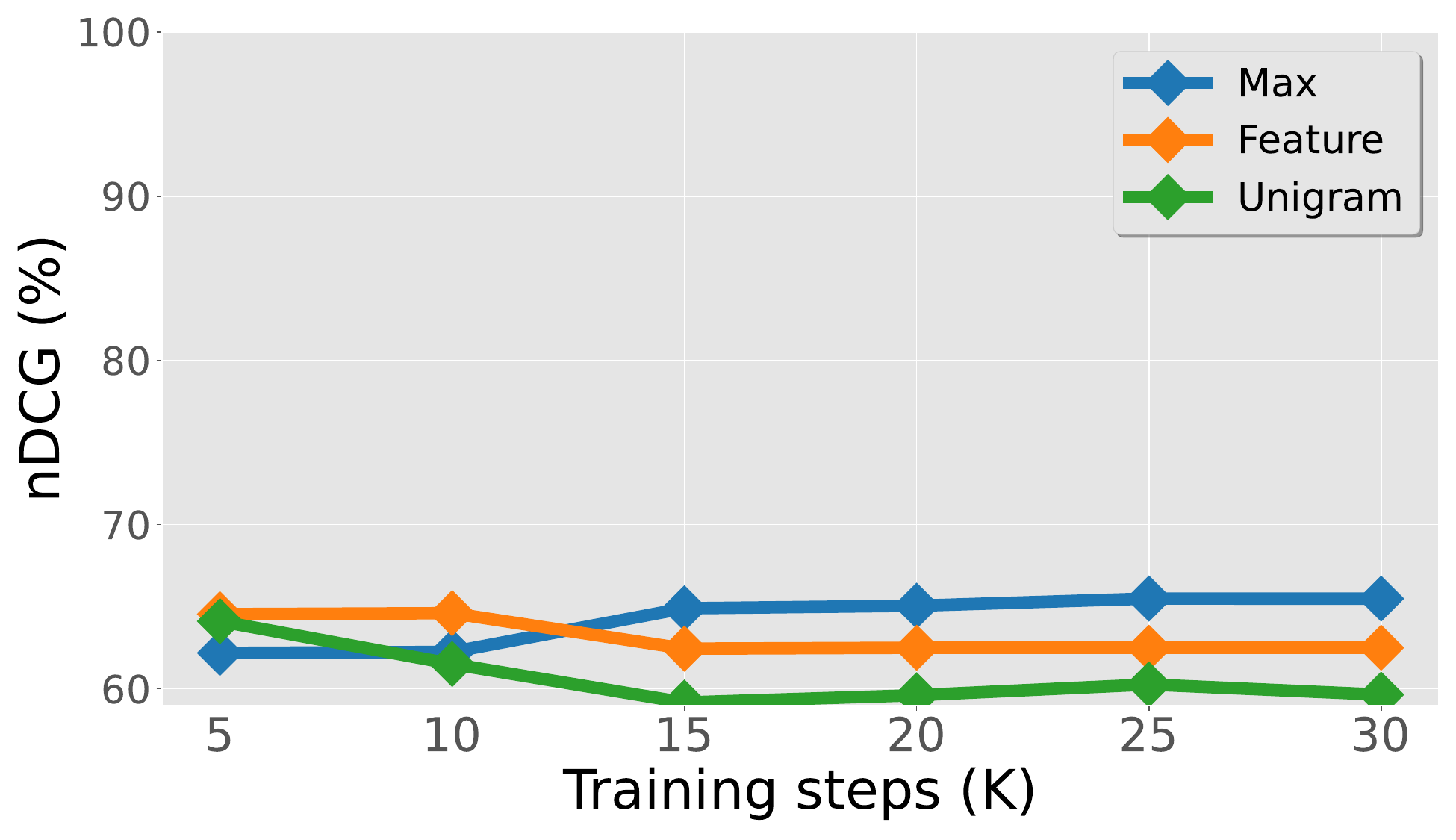}
        \caption{COPA}\label{ndcg_copa}
    \end{subfigure}%
    \begin{subfigure}[b]{0.44\linewidth}
        \centering
        \includegraphics[width=\linewidth]{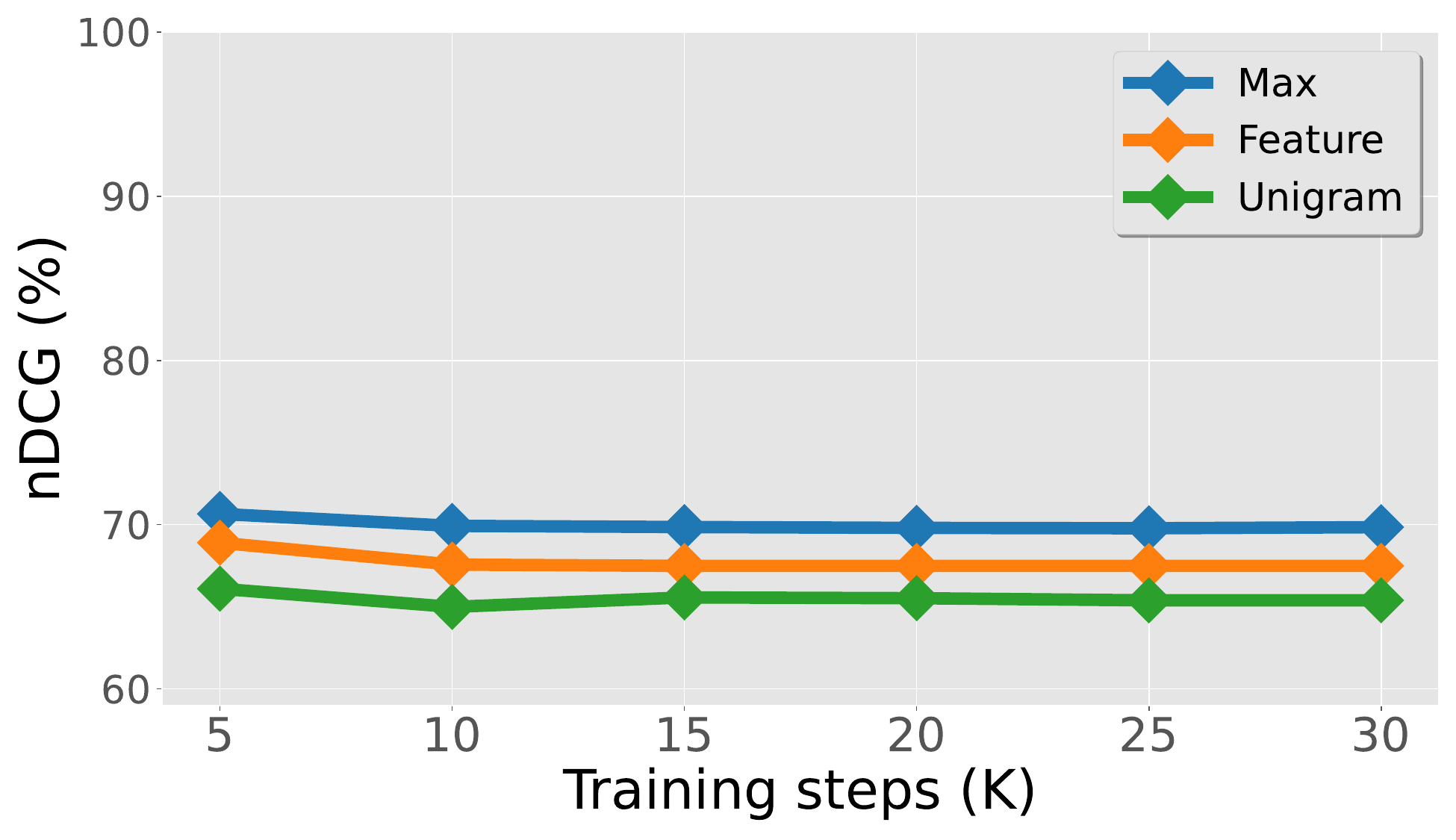}
        \caption{CB}\label{ndcg_cb}
    \end{subfigure}%
    \caption{Comparison of task embedding construction methods on various training steps, with intervals of 5K. The x-axis denotes the training steps of prompt-tuning.}
    \label{fig:ndcg_ten_tasks}
\end{figure*}

Figure~\ref{fig:ndcg_ten_tasks} analyzes how training steps for prompt tuning affect ranking prediction across various task embedding constructions, \bssc{MAX}, \bssc{Feature} and \bssc{Unigram}.
We examined the prompt weights trained at intervals of 5K, up to 30K, using nDCG for ranking prediction.
Three construction methods of task embeddings were compared across ten downstream tasks, indexed alphabetically from \ssc{BoolQ} (\hyperref[ndcg_boolq]{a}) to \ssc{CB} (\hyperref[ndcg_boolq]{j}).

\paragraph{Tasks with very limited data exhibit low nDCG scores.}
We found that the three methods performed well on five tasks, showing high nDCG scores. For instance, in \ssc{BoolQ}, \ssc{STS-B}, \ssc{CR}, \ssc{MRPC}, and \ssc{WSC}, all three methods demonstrated similar performance with relatively flat performance curves. 

We further observed the significant variability in task prediction performance across four tasks: \ssc{CoLA}, \ssc{RTE}, \ssc{COPA}, and \ssc{CB}. 
Notably, \ssc{COPA} and \ssc{CB} presented considerable challenges due to their limited availability of labeled data. As a result, the computed nDCG scores for these tasks were notably lower compared to other downstream tasks, underscoring the difficulty in identifying effective intermediate tasks.

\paragraph{\bssc{MAX} yields superior performances in task prediction.}
Across 10 downstream tasks, we observed that \bssc{MAX} generally yields superior nDCG scores.
On \ssc{CoLa}, \ssc{RTE}, and \ssc{COPA}, nDCG surpasses \bssc{Feature} after 15K training steps.
For \ssc{CB}, \bssc{MAX} excels in capturing the essence between intermediate tasks during continual prompt tuning on challenging low-resource tasks.
This highlights the importance of measuring token-wise similarity between source and target prompts for improved performance.
Our analysis suggests that \bssc{MAX} method tends to perform better in certain scenarios, emphasizing its effectiveness in ranking prediction compared to other methods.

\paragraph{Longer training leads to better performance.}
Furthermore, \bssc{MAX} achieves higher task prediction performance with longer training steps. Furthermore, \bssc{MAX} achieves higher task prediction performance with longer training steps. For example, in tasks such as \ssc{CoLA}, \ssc{Wic}, and \ssc{RTE}, \bssc{MAX} shows marked improvements in the ranking prediction with extended training durations.

\end{document}